\documentclass[preprint,12pt]{elsarticle}

\usepackage{amssymb}
\usepackage{amsmath}
\usepackage{algorithm}
\usepackage{algpseudocode}
\journal{Expert Systems with Applications}

\begin{document}
\makeatletter
\def\ps@pprintTitle{%
 \let\@oddhead\@empty
 \let\@evenhead\@empty
 \let\@oddfoot\@empty
 \let\@evenfoot\@oddfoot}
\makeatother

\begin{frontmatter}



\title{Model-Free Surrogate-Assisted Neural Architecture Search for Evolving Variable-Length Dense Blocks}


\author[1,4]{Asif Ameer\corref{cor1}}
\ead{asif.ameer@nu.edu.pk}

\author[2]{Maryam Bashir}
\ead{maryam.basher@nu.edu.pk}

\author[2]{Irfan Younas}

\author[3]{Muhammad Fayyaz}
\ead{m.fayyaz@nu.edu.pk} 

\cortext[cor1]{Corresponding author}

\affiliation[1]{organization={Department of Artificial Intelligence \& Data Science, FAST National University of Computer \& Emerging Sciences (NUCES)},%
            addressline={Chiniot-Faisalabad (CFD) Campus}, 
            city={Chiniot},
            postcode={35400}, 
            state={Punjab},
            country={Pakistan}}

\affiliation[2]{organization={Department of Artificial Intelligence \& Data Science, FAST National University of Computer \& Emerging Sciences (NUCES)},%
            addressline={Lahore Campus}, 
            city={Lahore},
            postcode={54000}, 
            state={Punjab},
            country={Pakistan}}

\affiliation[3]{organization={Department of Computer Science, FAST National University of Computer \& Emerging Sciences (NUCES)},%
            addressline={Chiniot-Faisalabad (CFD) Campus}, 
            city={Chiniot},
            postcode={35400}, 
            state={Punjab},
            country={Pakistan}}

\affiliation[4]{organization={Department of Computer Science, FAST National University of Computer \& Emerging Sciences (NUCES)},%
            addressline={Lahore Campus}, 
            city={Lahore},
            postcode={54000}, 
            state={Punjab},
            country={Pakistan}}
\begin{abstract}
Neural Architecture Search (NAS) has emerged as a powerful paradigm for automatically designing deep neural networks; however, its practical adoption is often limited by substantial computational cost. To alleviate expensive full-training evaluations, surrogate-based methods have been introduced to estimate network performance efficiently. Nevertheless, existing approaches—particularly model-based surrogates—require training a large number of candidate architectures and involve additional optimization overhead. In this work, we propose a Model-Free Surrogate PSO Network (MFSPNet) for evolving convolutional neural network architectures. The proposed method integrates a lightweight model-free surrogate predictor within a particle swarm optimization (PSO) framework, eliminating the need for pre-trained surrogate models. Specifically, MFSPNet introduces two key contributions: (1) a validation-loss–driven exponential moving average estimator (VLE-EMA) that captures early generalization behavior for reliable architecture ranking; and (2) a block-based dense connection strategy that enables effective stacking of evolved blocks while mitigating vanishing-gradient issues. This design also facilitates transferability of learned blocks across datasets. Extensive experiments demonstrate that MFSPNet achieves competitive performance with significantly reduced computational cost. Under a consistent training protocol with ten independent runs, the proposed method attains error rates of 3.91\%, 17.68\%, and 1.91\% on CIFAR-10, CIFAR-100, and SVHN, respectively, along with top-1/top-5 error rates of 28.29\%/12.82\% on ImageNet, while requiring less than three GPU days for architecture search. Due to computational constraints, the ImageNet result is based on a single run and should be interpreted as indicative of scalability. Overall, MFSPNet provides an efficient and reliable framework for cost-aware neural architecture search.
\end{abstract}

\begin{graphicalabstract}
\begin{figure}[h]%
\centering
\includegraphics[width=1.0\textwidth]{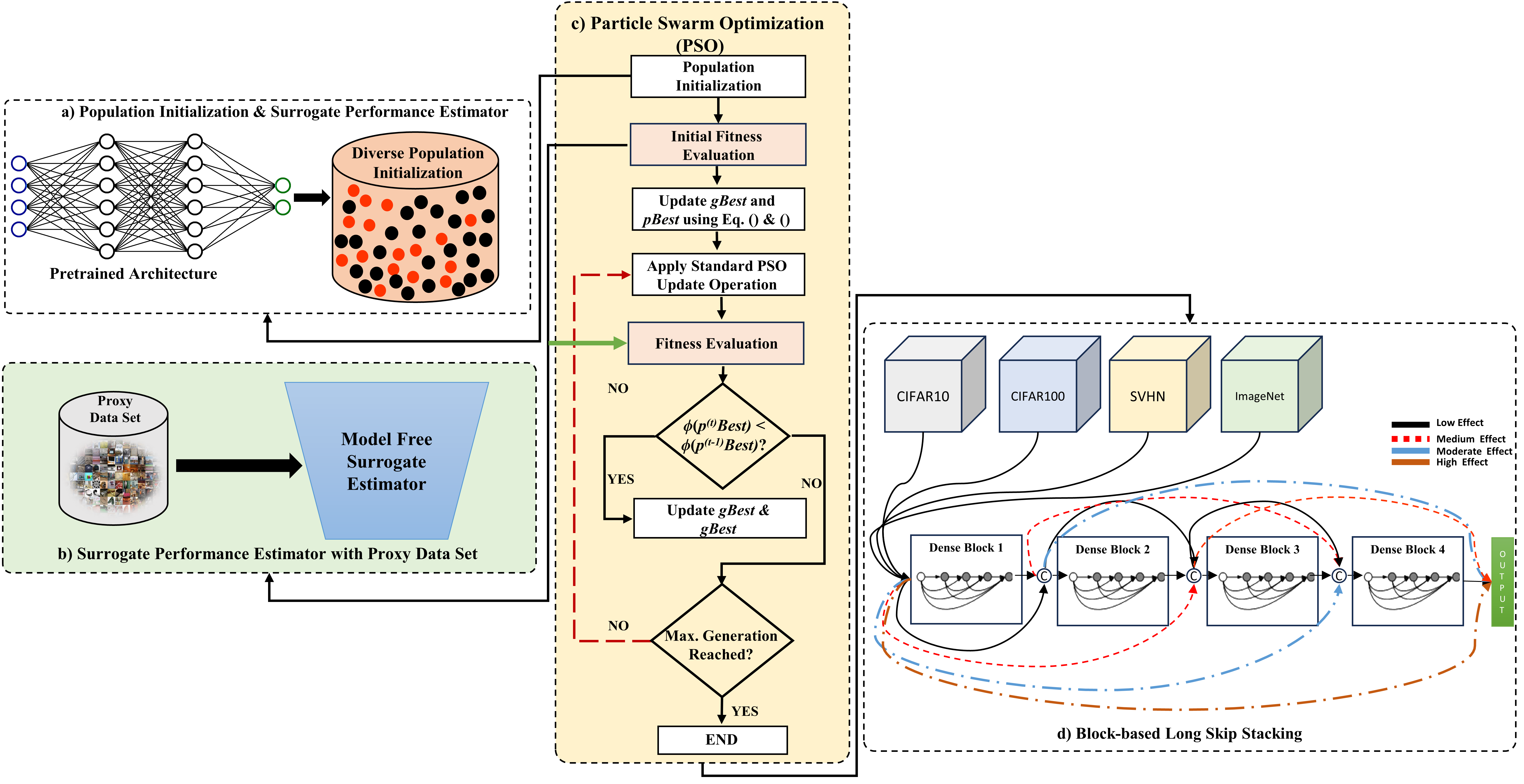}
\end{figure}\end{graphicalabstract}

\begin{highlights}
\item A PSO-based strategy to effectively search the dense block for image classification.

\item A model-free surrogate performance predictor to estimate for fitness evaluation.

\item A surrogate dataset estimates validation loss as a fitness function during evolution.

\item The efficacy of evolved block is analyzed on different large-scale dataset. 

\item A stacking strategy with block-based skip-connections is introduced.
\end{highlights}

%% Keywords

\end{frontmatter}


%% INTRODUCTION %%
\section{Introduction}\label{sec1-introduction}
Deep learning (DL) has significantly advanced various fields, including language translation \cite{wu2016google}, computer vision \cite{He2015}, and healthcare \cite{abramoff2016improved}. In the realm of image classification, convolutional neural networks (CNNs) have achieved remarkable success in recent years, consistently surpassing previous benchmarks for classification accuracy \cite{wang2021surrogate, zhao2019object}.

Advancements in DL architectures are often driven by the introduction of components such as convolutions, skip-connections, and normalization techniques \cite{lecun1998gradient, huang2017densely}. These innovations have enabled the replacement of shallow models with manually crafted features by deeper, larger, and more robust neural networks capable of extracting meaningful features directly from raw data. A prominent trend in improving classification error has been the increase in the number of hidden layers in CNNs. For instance, AlexNet \cite{real2017large} utilized only five convolutional layers, whereas VGGNet \cite{simonyan2014very} includes tens of layers, and architectures such as ResNet \cite{He2015} and DenseNet \cite{huang2017densely} contain hundreds of layers.

Recent advancements also highlight the integration of skip-connections, as seen in architectures such as ResNet, DenseNet, wide residual networks \cite{Zagoruyko2016}, and PyramidNets \cite{han2017deep}, where non-adjacent layers are connected through shortcut pathways. These connections have transformed the conventional feed-forward CNN architecture, enabling more flexible and effective network designs. However, while structural innovations such as batch normalization and skip-connections have led to significant improvements across various domains, they also pose considerable challenges. Training deeper and more complex architectures remains difficult due to the absence of consistent foundational principles for training novel designs \cite{huang2017densely}. As a result, neural network training often relies on extensive hyperparameter searches, which are computationally expensive and involve complex optimization strategies. Additionally, the precise impact of architectural modifications on network performance is not always clear. For example, while batch normalization and skip-connections may enhance the training process, their contribution to actual performance improvement is sometimes ambiguous \cite{wang2019evolving}. This has fueled significant interest in the automatic discovery of optimal DL architectural parameters.

Within this field, evolutionary computation (EC) and reinforcement learning (RL) have emerged as popular approaches. Studies \cite{Zoph2017, brock2017smash, zoph2018learning} demonstrate that RL-based methods can produce CNNs that outperform manually designed architectures. Similarly, EC-based approaches for evolving CNNs have shown promising results \cite{Xie2017, wang2021surrogate, real2019regularized}. Despite these advancements, many methods achieve high-performing CNNs only after evaluating a vast number of architectures, incurring substantial computational costs. For instance, Zoph et al. and Lee et al. reported in \cite{zhong2018practical, Lee2014} that achieving state-of-the-art CNNs required 22,400 GPU-days and 2,000 GPU-days, respectively. Such approaches often involve parallel trials on hundreds of GPUs, making them inaccessible to most researchers and practitioners due to limited computational resources.

To address this challenge, this article proposes the use of a model-free surrogate performance predictor (i.e., a training-less approach) to assist the evolutionary process. Reducing search cost is not only a methodological contribution but also of practical importance in scenarios where computational resources are limited, such as UAV-based monitoring, agricultural imaging, or embedded vision systems. By lowering the GPU hours required for architecture search, our method increases accessibility to practitioners without high-end hardware. This methodology aims to significantly reduce computational costs while enabling the discovery of high-performing CNN architectures in a more resource-efficient manner.

Many NAS methods, including AmoebaNet \cite{real2019regularized}, introduce architectural innovations (e.g., various cell patterns, mutation or crossover operators) to trade off accuracy and efficiency. In particular, EffPNet \cite{wang2021surrogate} employs a model-based surrogate, trained on a large dataset of fully trained architectures, to classify performance. In contrast, MFSPNet employs a model-free surrogate estimator that requires no prior surrogate training and considers validation loss dynamics, giving higher weight to later epochs. This eliminates the need for large datasets of trained architectures, thereby improving search efficiency instead of relying solely on architecture-level innovations.

The proposed approach draws inspiration from DenseNet \cite{huang2017densely}, EffPNet \cite{wang2021surrogate}, and TSE-EMA \cite{ru2021speedy}. EffPNet introduced a method for efficiently searching CNN architectures by employing Particle Swarm Optimization (PSO) to evolve optimal individual blocks rather than entire architectures. This strategy significantly reduces computational cost since training a single block requires far fewer resources than training an entire CNN, thereby accelerating the search process. Additionally, EffPNet demonstrated the applicability of transferring the best blocks learned from a proxy dataset to similar datasets in different domains. This paper adopts the strategy of evolving individual blocks and investigates the transferability of the evolved block.

DenseNet \cite{huang2017densely} is another key inspiration. By incorporating skip-connections between layers, DenseNet effectively enhances performance. However, its fixed hyperparameter—the growth rate—assigned uniformly to every composite layer within a dense block may not always yield optimal results. To address this, the proposed approach integrates an evolutionary computation technique to explore varying growth rates for each layer, enabling more flexible optimization. PSO is selected due to its simplicity, computational efficiency, and effectiveness in optimizing diverse functions \cite{schutte2005particle}.

Furthermore, the proposed method incorporates a simple model-free surrogate performance predictor. This predictor provides a computationally inexpensive estimate of the generalization performance ranking, serving as a substitute for costly function evaluations during the search process. This integration aims to streamline optimization while maintaining high performance.

\subsection{Goal}
The objective of this research is to evaluate the effectiveness of a model-free surrogate performance predictor in assisting an evolutionary computation method to efficiently evolve dense blocks. This approach aims to eliminate the need to train large numbers of architectures, which is typically required for building surrogate models. To achieve this objective, the following tasks are performed:

\begin{itemize}
 \item A model-free surrogate performance predictor is proposed to estimate the performance of candidate architectures, such as CNNs. The surrogate evaluates candidate architectures by summing validation losses obtained through training for a limited number of epochs using the SGD optimizer. This approach is inspired by theoretical studies linking training speed estimates to generalization \cite{ru2021speedy, hardt2016train}.

 \item To validate the efficacy of the proposed methodology, an in-depth analysis of the surrogate performance predictor is conducted. To accelerate the evolutionary process, the surrogate predictor eliminates underperforming candidate architectures after a few epochs of SGD training. Furthermore, the evolutionary process is analyzed and visualized to provide insights into convergence behavior.

 \item Finally, the block stacking process is executed, and the transferability of the evolved dense block is assessed on other datasets. To enhance the depth of the DenseNet architecture, the stacking process leverages a block-based skip-connection strategy.
\end{itemize}

\subsection{Abbreviations}
The list of abbreviations used in this paper is provided in Table~\ref{tab:notations}.

\begin{table}[ht]
\centering
\caption{List of notations used in this paper.}
\label{tab:notations}
\begin{tabular}{p{0.2\linewidth} p{0.6\linewidth}}
\hline
\textbf{Notation} & \textbf{Description} \\
\hline
$k_j$                  & Growth rate at layer $j$ \\
$\xi_i$                & Position of the $i_{\text{th}}$ particle of PSO \\
${\upsilon}_i$         & Velocity of the $i_{\text{th}}$ particle of PSO \\
$\phi$                 & Objective function \\
$p_{best_{i}}$         & Personal best of particle $i$ \\
${g}_{\text{best}}$    & Global best particle \\
${\omega}$             & Inertia weight \\
$c_1$                  & Cognitive weight \\
$c_2$                  & Social weight \\
${\rho}_1, {\rho}_2$   & Random numbers (uniform distribution) \\
$\gamma$               & Loss decay factor \\
$H_l$                  & Output of the $l_{\text{th}}$ dense block \\
$\psi$                 & Block-to-block skip-connection weight decay \\
\hline
\end{tabular}
\end{table}

The structure of this article is as follows: Section \ref{sec2:RL} provides the background and literature review, while Section \ref{sec3:methodolgy} outlines the details of the proposed method. The design of experiments is described in Section \ref{sec4:Experiments}, followed by the presentation and analysis of results in Section \ref{sec5:Results}. Finally, Section \ref{sec6:Conclusion and Future Work} concludes the study and discusses potential future research directions. The notations listed in Table \ref{tab:notations} are used throughout the paper to facilitate the understanding of the mathematical formulations and algorithms presented in subsequent sections.

\section{Background and Preliminaries}\label{sec2:RL}
\subsection{Dense Block Architecture}\label{sec2:Dense Block Architecture}
Since this article will employ the dense block, which is considered the most basic component in DenseNet \cite{huang2017densely}, so its details are discussed. The dense block consisting of four composite layers is illustrated in Fig. \ref{fig-dense_block}. Starting from the left, the convolutional layer extracts a variety of feature maps as input \cite{wang2019evolving}. After receiving the input, the first composite layer generates several output feature maps. In some CNN architectures, the input of all the following layers in a dense block is produced by concatenating the output feature maps of the current layer with the input of itself. The subsequent layers also follow the same process. A dense block functions by concatenating the output feature maps of all preceding layers with the input of the current layer. This means the input to the first layer (layer 0) is the original input feature maps of the dense block. Subsequently, the input to layer $l$ is formed by concatenating the output feature maps from layers 0 through $l-1$. Finally, the output of the dense block is generated by concatenating all the preceding layers output in the block, along with the original input feature maps. Layers within one dense block are composite, consisting of three layers: batch normalization (BN), rectified linear unit (ReLU), and convolutional with 3x3 filters \cite{huang2017densely}. In Fig. \ref{fig-dense_block}, each composite layer comprises of BN-ReLU-Conv layers. The input size at $(l + 1)_{th}$ layer can be calculated with Eq. \ref{Eq_growth},

\begin{equation}\label{Eq_growth}
N_{l+1}=\sum_{j=0}^l k_j
\end{equation}
where $k_0$ represents the input i.e (feature maps) of dense block and $k_l$ represents the $l_{th}$ layer output. In DenseNet architecture \cite{huang2017densely} $k_i$ is considered as a growth rate hyperparameter, which plays a pivotal role in network learning process. This growth rate refers to the total feature maps generated by every layer in a dense block. In the original DenseNet paper, this growth rate parameter is fixed and denoted with  $r$  for an entire dense block. Which can be calculated by transforming Eq \ref{Eq_growth} as $N_{l+1}= k_0 + r * l$:
\begin{figure}[h]%
\centering
\includegraphics[width=0.9\textwidth]{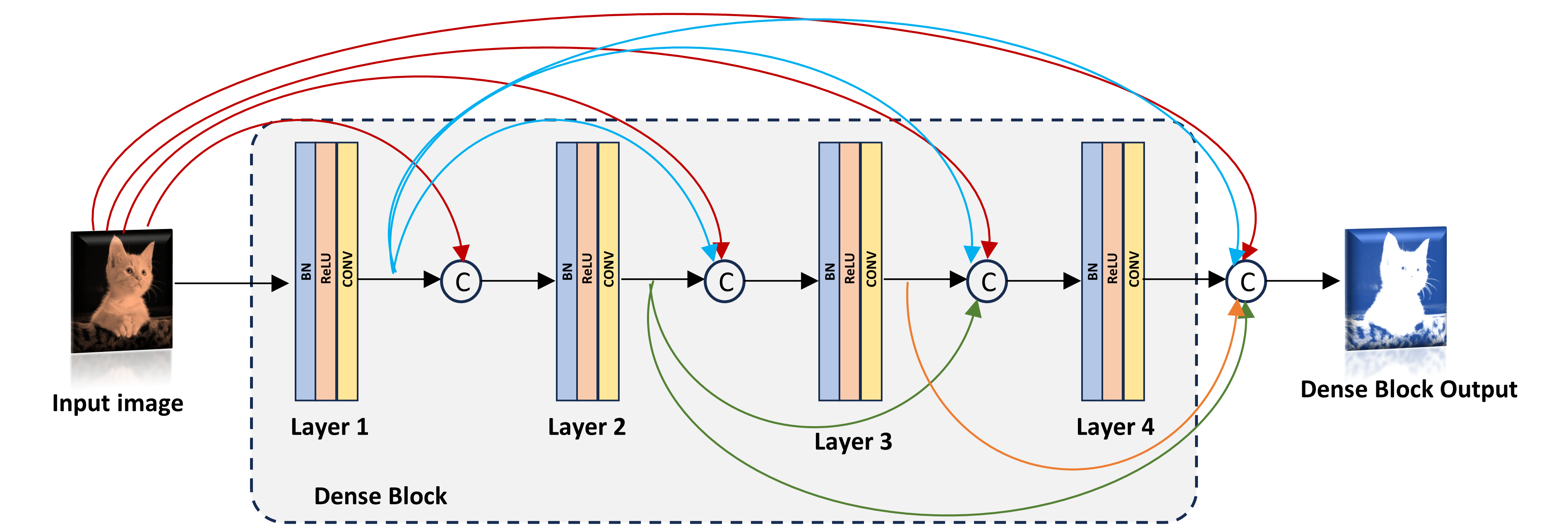}
\caption{Architecture of a dense block with four composite layers and skip-connections. 
Each layer reuses features from all preceding layers, improving gradient flow and learning.}\label{fig-dense_block}
\end{figure}

\subsection{Particle Swarm Optimization (PSO)}\label{sec2:Particle Swarm Optimization}
The fundamental ideas of PSO include behaviors like fish schooling, bird flocking, and swarm theory \cite{van2006study}. A swarm of particles continuously changes their relative locations from one iteration to another, enhancing the performance of the PSO algorithm in carrying out the search process effectively \cite{gad2022particle}. To get an optimal solution, each particle in the swarm improves towards its own personal best $\mathbf{p}_{best}$ position and toward the global best $\boldsymbol{g}_{best}$ position. For the minimization problem, the personal best and global best positions are defined as follows \cite{kennedy1995particle, eberhart1995new}:

\begin{equation}
\mathbf{p}_{best_{i}}^{(t)} = \boldsymbol{\xi}_i^* \mid \phi(\boldsymbol{\xi}_i^*) = \min_{k=1,2,\ldots,t} \{ \phi(\boldsymbol{\xi}_i^k) \}, \quad \text{for} \quad i \in \{1, 2, \ldots, N\}.
\end{equation}
\begin{equation}
\boldsymbol{g}_{\text{best}}^t = \boldsymbol{\xi}_*^t \mid \phi\left(\boldsymbol{\xi}_*^t\right) = \min_{\substack{i=1,2,\ldots,N \\ k=1,2,\ldots,t}}\left(\left\{\phi\left(\boldsymbol{\xi}_i^k\right)\right\}\right)
\end{equation}
Where $i$ represents the index of a particle, $t$ represents the iteration number. The fitness function, to be optimized, is represented by $\phi$. $\boldsymbol{\xi}$ denotes the position vector or potential solution, and $N$ represents the population size.
The velocity and position of each particle $i$ are updated in each generation using Eqs.~\ref{eq_velocity} and \ref{eq_position}, respectively, as originally introduced by Kennedy and Eberhart~\cite{kennedy1995particle,eberhart1995new}:

%%% PSO EQUATIONS %%%
\begin{equation} \label{eq_velocity}
\upsilon_{i}^d(t+1) = \omega \upsilon_{i}^d(t) + c_1 \rho_1\left(p_{i}^d(t) - \xi_{i}^d(t)\right) + c_2 \rho_2\left(g_i^d(t) - \xi_{i}^d(t)\right)
\end{equation}
\begin{equation}\label{eq_position}
\xi_{i}^d(t+1) = \xi_{i}^d(t) + \upsilon_{i}^d(t+1) \text{.}
\end{equation}
Where $\xi$ and $\upsilon$ denote the position and velocity the the PSO particle, respectively. During position and velocity update by Eq. \ref{eq_position} and \ref{eq_velocity}, $\xi^d$ and $\upsilon_{i}^d(t+1)$ represent the position and velocity of the particle at $d_{th}$ dimension. The $t$ and $(t+1)$ represents the position or velocity of current and next iteration respectively. In addition to this, the velocity update, equation \ref{eq_velocity} includes inertia, cognitive and social weight represented with $\omega$, $c_1$ and $c_2$. $\rho_1$ and $\rho_2$ are the randomly generated numbers uniformly distributed in range $[0, 1]^d$. 

Although several other population-based optimization strategies such as Genetic Algorithms (GA), Ant Colony Optimization (ACO), and Differential Evolution (DE) have been explored in NAS, PSO is particularly well suited for the surrogate-assisted block search proposed in this work. GA relies on crossover and mutation operations that become increasingly complex when handling variable-length and structured CNN block encodings \cite{maldini2024review}. ACO, while effective for discrete combinatorial problems, requires pheromone matrix updates whose computational and memory overhead grow rapidly with search-space dimensionality. DE requires careful tuning of mutation and crossover strategies, and its performance is known to be sensitive to population size and control parameters, particularly in high-dimensional and structured search spaces such as variable-length CNN block representations \cite{das2011differential}. This sensitivity can increase the number of function evaluations needed to maintain adequate diversity during the search process. Because surrogate estimates can be noisy during early generations, PSO’s collective information sharing between particles helps stabilize convergence while still preserving exploration. Furthermore, PSO requires fewer control parameters and exhibits fast convergence, making it computationally attractive when each candidate evaluation involves partial network training. These characteristics make PSO well aligned with the objectives of MFSPNet: reducing search cost while maintaining reliable architecture optimization \cite{chauhan2025learning}.

\subsection{Surrogate Models in NAS}\label{sec2:Surrogate Performance Predictors}
The use of surrogate-assisted techniques in evolutionary optimization problems is not new.  This was developed to address the practical difficulties encountered in engineering optimization problems when there are no analytical solutions to objective functions or the fitness evaluation i.e. (performance of candidate architecture) is time-consuming, taking many hours or even days \cite{liu2021survey}. When employing population-based optimization techniques in NAS, such as Evolutionary Algorithms (EAs), to find the best candidate solution for a particular job, a number of potential solutions must be evaluated. Therefore, the expense of doing the computation in these particular circumstances is not feasible. To address this issue, surrogate-assisted methods explore the use of training surrogate models with a small data set. These models often require sufficient training, on pre-trained architectures, throughout the optimization process. Because model-free surrogate predictors frequently do not require explicit training on pre-trained candidate solutions, using them can also lessen the need for pre-trained architectures. A validation loss landscape-based proxy estimator is suggested as a surrogate predictor, to estimate the change in validation loss corresponding to architectural changes. An optimum neural architecture may be searched with a minimization function of validation loss. \cite{li2020neural}.

\section{Related Work}\label{sec2:Related Work}
Recent studies have concentrated on reinforcement learning (RL) methodologies for the automated construction of convolutional neural networks (CNNs), using the results of initial Neural Architecture Search (NAS) frameworks \cite{zoph2016neural}. Reinforcement Learning-based Neural Architecture Search utilizes a controller network, commonly an RNN or policy-gradient agent, in order to build candidate architectures and obtain performance-driven rewards. Furthermore, NAS \cite{zoph2016neural}, NASNet \cite{zoph2018learning}, and ENAS \cite{pham2018enas} proved the possibility of learning architectural design policies directly from validation input. Despite attaining state-of-the-art performance, these techniques demanded considerable computing power, utilizing as much as 22,400 GPU-days for NAS \cite{zoph2016neural}. Afterward improvements, like AmoebaNet \cite{real2019regularized}, enhanced search consistency by combining regularization and evolutionary search; however, the computing expense remained too high. Despite their popularity, RL-based approaches usually suffer both slow convergence and high sample inefficiency, since each design has to be trained or partially trained to generate a reward. Moreover, controller networks themselves demand accurate hyperparameter tuning, that additionally contributes to the computational burden. Therefore, the focus of research advanced towards Evolutionary Computation (EC)-based NAS approaches that attempted to preserve the versatility of population-based search.

The LS-Evolution technique was proposed in an early study that uses evolutionary algorithms to evolve CNNs, which demonstrated the promising performance at remarkable computational expense. Later, a few effective EC-based techniques were presented, including GeNet \cite{xie2017genetic}, CGP-CNN \cite{suganuma2017genetic}, and EIGEN \cite{ren2019eigen}, that lower computing costs by compromising classification accuracy in comparison to LS-Evolution. The AmoebaNet \cite{real2019regularized} and AECNN \cite{sun2019completely} were also proposed by modifying the search strategies. To search for the best CNN architecture, AmoebaNet suggested a regularized evolutionary algorithm results improved performance due to reduced candidate sampling. On the other hand, AECNN expedited the search process by analyzing a smaller number of CNNs, using less computational expense; nevertheless, in contrast to AmoebaNet, the classification accuracy degraded. Moreover, EffPNet in \cite{wang2021surrogate} also alleviated the burden of an evolutionary algorithm by employing the classification surrogate model to assist the searching process. This approach evolved the dense block architecture and achieved the best classification accuracy with minimal GPU usage. Additionally, gradient-based NAS techniques such as ProxylessNAS \cite{cai2019proxylessnas} and DARTS \cite{liu2019darts} have been proposed to enhance search performance by reducing the discrete architectural space into a differentiable form, which enabled end-to-end optimization less resource intensive. While these techniques offer quicker convergence, but typically indicate instability or performance loss across datasets owing to approximation errors in gradient reduction. In contrast, our surrogate-assisted PSO approach preserves reliability by explicitly capturing the performance landscape with validation-based lightweight predictors, consequently minimizing the instability associated with gradient reduction. In this article, a novel and reliable lightweight model-free surrogate predictor is proposed to aid
the evolutionary process to reduce the computational burden during search. The sole purpose of choosing the EC technique for searching best neural architecture is its fast convergence and less computational complexity.

In addition to surrogate-assisted and gradient-based NAS methods, recent studies have introduced training-free (zero-cost) proxy metrics such as NASWOT \cite{Mellor2020NASwithoutTraining} and SynFlow \cite{Tanaka2020SynFlow}, which estimate architecture quality without performing SGD optimization. These approaches analyze properties of randomly initialized networks and therefore incur negligible computational cost. However, because they do not observe learning dynamics, their ranking consistency with final validation accuracy may degrade for deeper architectures, stacked block configurations, or cross-dataset transfer scenarios. In particular, zero-cost proxies rely on initialization-dependent signals and do not capture how feature representations evolve during training, which is critical when transferring architectures across datasets with different distributions (e.g., CIFAR to ImageNet). 

The proposed VLE-EMA differs fundamentally from such training-free proxies. Although it remains model-free (i.e., it does not rely on a learned regression or classification surrogate model), it performs a small amount of actual training (e.g., $c=15$ epochs) and leverages validation-loss dynamics as the performance signal. This enables the estimator to capture early generalization behavior, making it more robust for architecture ranking in transfer settings while maintaining a substantially lower computational cost than full training.

A comparative summary of representative NAS approaches in terms of computational cost (GPU-days), classification error, parameter size, and FLOPs is presented in Table~\ref{tab:TblComparisonCIFAR_10}. This comparison highlights the significant trade-off between performance and computational expense in existing RL-, EC-, and gradient-based NAS methods. In particular, many high-performing approaches require substantial GPU resources, which limits their practical applicability. Based on this observation, it is evident that automatically designing CNN architectures remains challenging due to the balance between computational efficiency and predictive performance. To address this limitation, this work proposes a novel model-free surrogate predictor to guide the evolutionary search process, reducing computational overhead while maintaining competitive accuracy. The use of an EC-based strategy is motivated by its fast convergence and lower computational complexity compared to alternative search paradigms~\cite{real2019regularized, wang2021surrogate}.

\section{Proposed Methodology}\label{sec3:methodolgy}
\begin{figure}[h]%
\centering
\includegraphics[width=1.0\textwidth]{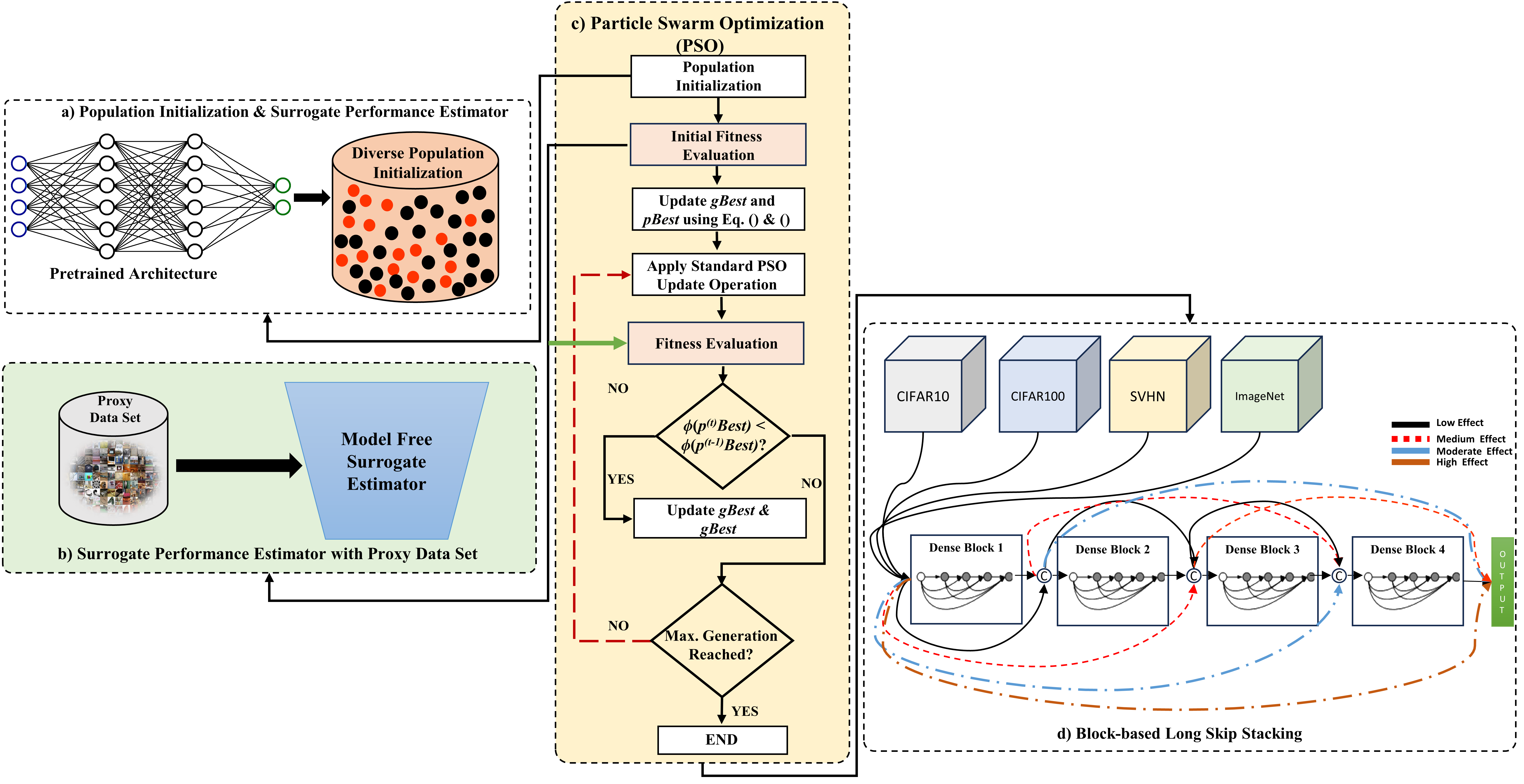}
\caption{Overview of the MFSPNet framework. (a) Population initialization of candidate dense blocks. (b) Low-resolution proxy dataset generation for surrogate-based fitness evaluation. (c) Surrogate-assisted PSO search to evolve the optimal dense block. (d) Block stacking and cross-dataset transfer evaluation on CIFAR-100, SVHN, and ImageNet.}\label{fig1:GA-EMA}
\end{figure}

This section exhibits the details of the suggested technique, known as MFSPNet (Model-free Surrogate PSO Network). The architectural framework of the proposed approach is depicted in Fig. \ref{fig1:GA-EMA}. Initially, to lessen the burden of the evolutionary search process, the proxy dataset is generated from the standard dataset by employing the strategy in Section \ref{sec3:Proxy Data Set} and illustrated in Fig. \ref{fig1:GA-EMA}(b). This proxy dataset is used to evaluate the fitness of each particle during the evolutionary process. A model-free surrogate predictor is used to evaluate the fitness of the block, rather than training the CNNs with expensive SGD during the evolutionary process, as shown in Fig. \ref{fig1:GA-EMA}(b)--(c). The complete process is described in Section \ref{sec3:Surrogate Performance Predictor}. 
The proposed MFSPNet framework greatly reduces the computational cost of evolutionary NAS by integrating a model-free surrogate performance estimator with a proxy dataset. Subsequently, the framework employs a block-based skip-connection stacking strategy to generate CNN architectures with varying architectural complexities, as illustrated in Fig. \ref{fig1:GA-EMA}(d) and described in Section \ref{sec3:Block Stacking}.
The block stacking process is evaluated using only the training set, and the final CNN is further evaluated on the training and validation datasets. To ensure a fair comparison with peer competitors, the classification accuracy on the test set is reported. The evolved block is then stacked and transferred to another domain.

\subsection{Particle Encoding}\label{sec3:Particle Encoding}
\begin{figure}[h]
\centering
\includegraphics[width=1 \textwidth]{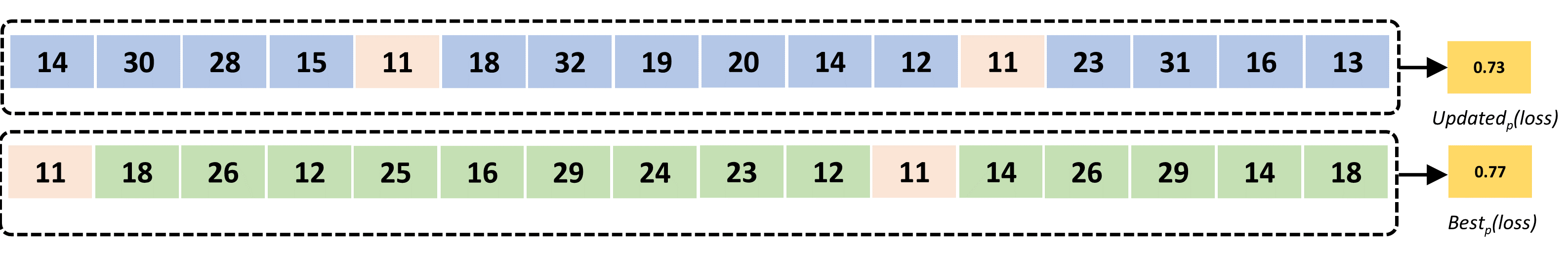}
\caption{Encoded candidate architecture for PSO evolutionary process. The maximum and minimum growth rate is set between 12-32. Fitness evaluation of two architectures during PSO is illustrated.}\label{fig-encoding}
\end{figure}

Searching for an entire CNN architecture directly leads to an extremely large and high-dimensional search space, which significantly slows convergence and increases the risk of unstable evolution. In addition, architectures discovered for one dataset often fail to generalize to others. To overcome these issues, we adopt a block-based search strategy where compact dense blocks are evolved as reusable building units. This reduces search complexity, improves evolutionary stability, and enables transferability across datasets by stacking the evolved blocks to form deeper architectures. The particle encoding method is exploited for encoding dense block hyperparameters \cite{huang2017densely}, which represent a variable number of composite layers within a block. In the DenseNet architecture, two hyperparameters influence the overall performance of the network—the number of composite layers within the block and their corresponding growth rates. In the DenseNet paper, a constant growth rate is suggested for all layers in the block. However, the fixed growth rate is not always optimal; thus, the proposed technique explores dense blocks with varied growth rates for each layer in DenseNet \cite{huang2017densely,li2021efficient}. Fig. \ref{fig-encoding} shows how a fixed-length vector is utilized to encode dense blocks with varying composite layer sizes as PSO particles.

The length of the vector represents the total number of layers in the dense block architecture, and a corresponding value of each cell denotes the growth rate of a specific layer of the block. A special flag value is used to disable a certain block layer, which introduces variability in the length of the block architecture. To convert the dense block to variable length, a flag value is also specified to disable the certain layer. Therefore, to encode the dense block architectures, three hyperparameters are specified and represented with PSO particles.

\begin{enumerate}
    \item The number of layers $(l_{max})$ in block architecture.
    \item The upper $(r_{upper})$ and lower $(r_{lower})$ ranges of growth rate at each layer.
    \item The special flag $(S_v)$ to represent the disabled layer.
\end{enumerate}

\begin{quote}
\textbf{Worked Example:}Flag based encoding for dense block.

Suppose: maximum layers $l = 8$, and 

growth rate range $[r_{lower}, r_{upper}] = [12, 32]$  

Step 1: Generate flags for each layer  
$\rightarrow [1, 0, 1, 1, 0, 1, 0, 1]$  

Step 2: Assign growth rates where flag = 1, and use 11 for disabled layers  
$\rightarrow [16, 11, 28, 20, 11, 24, 11, 12]$  

Final particle vector:  
\[
P = [16, 11, 28, 20, 11, 24, 11, 12]
\]

\textbf{Interpretation}: Layers 2, 5, and 7 are disabled (encoded by $11$). Active layers are assigned growth rates (16, 28, 20, 24, 12), yielding a dense block with 5 active layers of varying growth rates.
\end{quote}

Based on the experimental experience on DenseNet \cite{huang2017densely}, the ranges of growth rates $(r_{lower})$ and $(r_{upper})$ are adjusted between 12 and 32, respectively. A very small growth rate may adversely affect the output feature maps concatenation process. On the other hand, high ranges of growth rates may require extensive computational resources and lead to a slowdown of the search process. The special value $(r_{lower}-1 = 12 - 1 = 11)$ will be used as the layer-disabled flag. By employing the experience of \cite{wang2021surrogate}, the vector length is set to 16, which is entirely decided based on available hardware resources.  
The particle encoding process can be formally described using the pseudocode outlined in Algorithm \ref{alg:particle-encoding}.

\begin{algorithm}[H]
\caption{Particle Encoding Scheme for Dense Block}
\label{alg:particle-encoding}
\begin{algorithmic}[1]
\Require $l_{max}$: maximum number of layers in a block
\Require $r_{lower}, r_{upper}$: growth rate bounds
\Require $S_v$: special flag for disabling a layer
\Ensure Encoded particle vector $P$

\State Initialize empty vector $P$ of length $l_{max}$
\For{$i = 1$ to $l_{max}$}
    \State Randomly select flag $f \in \{0,1\}$ \Comment{$0$: disable, $1$: enable}
    \If{$f = 1$}
        \State Randomly assign growth rate $g \in [r_{lower}, r_{upper}]$
        \State $P[i] \gets g$
    \Else
        \State $P[i] \gets S_v$ \Comment{$S_v = r_{lower}-1$ denotes disabled layer}
    \EndIf
\EndFor
\State \Return $P$
\end{algorithmic}
\end{algorithm}

% \begin{algorithm}[ht]
% \caption{Particle Encoding Scheme for Dense Block}
% \label{alg:particle-encoding}
% \begin{algorithmic}[1]
% \Require $L$: maximum allowed layers in a block
% \Require $G_{\min}, G_{\max}$: minimum and maximum growth rates
% \Ensure Encoded particle vector $P$

% \State Initialize empty vector $P$ of length $L$
% \For{$i = 1$ to $L$}
%     \State Randomly select flag $f \in \{0,1\}$ \Comment{$0$: disable, $1$: enable}
%     \If{$f = 1$}
%         \State Randomly assign growth rate $g \in [G_{\min}, G_{\max}]$
%         \State $P[i] \gets g$
%     \Else
%         \State $P[i] \gets -1$ \Comment{$-1$ denotes disabled layer}
%     \EndIf
% \EndFor
% \State \Return $P$
% \end{algorithmic}
% \end{algorithm}

\subsection{Population Initialization}\label{sec3:Population Initialization}
A well-designed initialization strategy can greatly influence the diversity of the initial population and subsequently impact the convergence and exploration capabilities of the evolutionary NAS strategy \cite{liu2021survey}. In this study, we propose a novel particle initialization strategy that combines \textit{Latin hypercube sampling (LHS)} with a \textit{standard deviation} (SD) method to enhance the diversity of the initial population. The population is initialized with the procedure illustrated with Algorithm \ref{alg:population_initialization}. The suggested particle initialization technique uses the integrated benefits of LHS and the maximizing SD to build a diverse and representative population for PSO. By employing this diverse population initializing strategy, our technique boosts the exploration capabilities of the evolutionary search strategy and improves its convergence toward global optimum during search.

\begin{algorithm}
\caption{Population Initialization with LHS and SD Method}
\label{alg:population_initialization}
\begin{algorithmic}[1]
    \State \textbf{Input:} $n$ (number of particles), $d$ (dimension of search space)
    \State \textbf{Output:} Initial population $P$
    \State Initialize $P \leftarrow \emptyset$
    \State Use Latin Hypercube Sampling (LHS) to generate $n$ samples $S = \{ \mathbf{s}_1, \mathbf{s}_2, \ldots, \mathbf{s}_n \}$ in $d$-dimensional space
    \For{each sample $s_i \in S$}
    \State Compute the mean across dimensions:
    \[
    \bar{s}_i = \frac{1}{d} \sum_{j=1}^{d} s_{i,j}
    \]
    \State Compute the standard deviation across dimensions:
    \[
    \sigma_i = \sqrt{ \frac{1}{d-1} \sum_{j=1}^{d} \left( s_{i,j} - \bar{s}_i \right)^2 }
    \]
\EndFor
\State Sort $S$ in descending order of $\sigma_i$ and denote the reordered set by $S'$
    \For {each sample $\mathbf{s}_i \in S'$}
        \State Add sample $\mathbf{s}_i$ to population $P$
    \EndFor
    \State \textbf{Return} $P$
\end{algorithmic}
\end{algorithm}

% \begin{algorithm}[H]
% \caption{Population Initialization with LHS and SD Method}
% \label{alg_init_population}
% \begin{algorithmic}[1]
% \State \textbf{Input:} $n$ (number of particles), $d$ (dimension of search space)
% \State \textbf{Output:} Initial population $P$
% \State Initialize $P \leftarrow \emptyset$
% \State Use Latin Hypercube Sampling (LHS) to generate $n$ samples in $d$-dimensional space
% \State Calculate the standard deviation (SD) for each dimension across all samples
% \State Sort the samples based on the calculated SD in descending order
% \For{each sample $s$ in the sorted list}
%     \State Add sample $s$ to population $P$
% \EndFor
% \State \textbf{Return} $P$
% \end{algorithmic}
% \end{algorithm}

\subsection{Surrogate Performance Predictor}\label{sec3:Surrogate Performance Predictor}
A major challenge in surrogate-assisted NAS is estimating final performance without fully training each candidate architecture. Model-based surrogates require extensive fully evaluated networks and careful tuning, making them computationally expensive. Training-based proxies such as TSE rely on early training loss signals, which may correlate poorly with generalization performance. Uniform averaging of early losses can also obscure informative convergence trends. To overcome these issues, we adopt and adapt the TSE concept into a validation-loss driven exponentially weighted surrogate for more reliable architecture ranking \cite{baker2017accelerating, domhan2015speeding, klein2022learning}. However, training the model-based surrogate needs hundreds of completely evaluated architectures and optimizing its hyperparameters to reach sufficient extrapolation performance. We draw inspiration from the training speed estimator (TSE) and estimator with exponential moving average (TSE-EMA) represented with Eqs. \ref{Eq_TSE} and \ref{Eq_TSE-EMA}, respectively \cite{ru2021speedy}.\\

\textbf{Definition 1:} (Training Speed Estimator).\textit{Assume the loss function of neural network is represented with $\ell$. The $f_{\theta}(x)$ denotes the output of a neural network with input $x$ and trainable parameters $\theta$. After training the network $f(x)$ for $t$ epochs and $i$ minibatches of stochastic gradient descent (SGD) are denoted as $\theta_{t, i}$. Once the training is completed for $T$ epochs the training loss is calculated with Training Speed Estimate (TSE).}\cite{ru2021speedy}

\begin{equation}\label{Eq_TSE}
\mathrm{TSE}=\sum_{t=1}^T\left[\frac{1}{B} \sum_{i=1}^B \ell\left(f_{\theta_{t, i}}\left(\mathbf{X}_i\right), \mathbf{y}_i\right)\right]
\end{equation}
where mini-batch $x_{i}, y_{i}$ training loss is represented with $\ell$ at epoch $t$ and $B$ (batches) is the number of training steps within an epoch $t$.
% The second estimator in Equation \ref{} hypothesise that an estimator of network training speed that assigns higher weights to later epochs may exhibit a better correlation with the true generalisation performance of the final trained network. On the other hand, it is common for neural networks to overfit on their training data and reach near-zero loss after sufficient optimisation steps, so attempting to measure training speed solely based on the epochs near the end of training will be difficult and likely suffer degraded performance on model selection\cite{lewkowycz2020large}.
% \begin{equation}
% \text { TSE-EMA }=\sum_{t=1}^T \gamma^{T-t}\left[\frac{1}{B} \sum_{i=1}^B \ell\left(f_{\theta_{t, i}}\left(\mathbf{X}_i\right), \mathbf{y}_i\right)\right]
% \end{equation}

In addition to the original TSE formulation in Eq.~\ref{Eq_TSE}, Ru et al.~\cite{ru2021speedy} also introduced a variant based on exponential moving average of training losses (TSE-EMA), defined as:

\begin{equation}\label{Eq_TSE-EMA}
\mathrm{TSE\text{-}EMA}=\sum_{t=1}^T \gamma^{T-t}\left[\frac{1}{B} \sum_{i=1}^B \ell\left(f_{\theta_{t, i}}\left(\mathbf{X}_i\right), \mathbf{y}_i\right)\right]
\end{equation}

This formulation corresponds to the exponential moving average variant of TSE as presented in \cite{ru2021speedy}, where the estimator operates on training-loss trajectories to approximate optimization speed.

In this research, a model-free validation loss-based surrogate predictor is proposed to estimate the performance of the CNN block by employing the validation loss by extending the capabilities of training-speed estimator referred in Eq. \ref{Eq_TSE-EMA}. Initially, the dense block architecture represented by the particle is trained for \textit{validation-cutoff epochs} which does not require the huge number of pre-trained architectures to train model-based surrogate estimators. Secondly, such model-free training predictors also provide a good performance estimate by alleviating the need for expensive function evaluations during the search process  \cite{yang2023revisiting}. The shortcomings of TSE provided in Eq. \ref{Eq_TSE} assign equal weights to all the losses acquired during training epochs. However, our proposed surrogate predictor is not meant to predict the training speed, but the performance of the candidate architecture, represented with particle, on validation losses, will be analyzed with few training epochs. This predictor will give more importance to the validation losses acquired near the end of \textit{validation-cutoff epochs} as it is suggested that the neural network training as validation loss dynamics during the very early epochs are often unstable and do not necessarily reveal characteristics of the converged networks \cite{ru2021speedy}. 

Unlike TSE and its EMA-based variant (TSE-EMA), which operate on training-loss trajectories to estimate optimization speed, the proposed VLE-EMA applies the same exponential weighting mechanism to validation-loss signals.
Formally, the distinction lies in the substitution of the loss signal:
\[
\ell(\cdot) \rightarrow \ell_v(\cdot),
\]
which shifts the estimator from training dynamics to validation-based generalization behavior.
By applying EMA to validation losses, VLE-EMA emphasizes stable late-epoch generalization signals while suppressing early training noise. This shift from optimization-based estimation to generalization-aware estimation fundamentally changes the surrogate’s behavior, guiding the search toward architectures with robust validation performance rather than merely fast training convergence.
The underlying signal–noise model and mathematical justification distinguishing VLE-EMA from TSE-EMA are provided in Appendix A.

This modified model-free surrogate performance predictor is represented with Eq. \ref{performance_estimator}.
%% TSE Estimator with Penalizing Training Loss over Epochs %%
\begin{equation}\label{performance_estimator}
\text { VLE-EMA }=\sum_{t=1}^T \gamma^{T-t}\left[\frac{1}{{B}_{v}} \sum_{i=1}^{{B}_{v}} \ell_{v}\left(f_{\theta_{t, i}^v}\left(\mathbf{X}_i\right), \mathbf{y}_i\right)\right]
\end{equation}
where $\ell_v$ is the validation loss of a mini-batch $x_{i}, y_{i}$ at epoch t and $B_v$ (batches) is the number of validation steps within an epoch $t$.
% Our suggested model-free surrogate performance predictor, VLE-EMA, uses an exponential moving average of the total validation with $\gamma$ = 0.78, giving more weight to the sum of losses at later epochs than the information from the early training trajectories. 
The value of the exponential decay factor \(\gamma = 0.78\) was empirically determined through sensitivity analysis across multiple runs by varying \(\gamma \in [0.6, 0.9]\). This choice aligns with prior studies that employ exponential moving averages for stabilizing optimization dynamics, where decay factors in a similar range have been shown effective \cite{kingma2015adam,loshchilov2019adamw,polyak1992acceleration,wortsman2022soups}.

This value consistently provided the best correlation between early validation loss trends and final model accuracy, balancing responsiveness and stability in the surrogate estimation. The VLE-EMA surrogate estimator will compare the performance of the candidate solution and if the updated solution outperforms the former one, the solution will be updated for the next EC generation.

\subsection{Proxy Data Set}\label{sec3:Proxy Data Set}
To perform the evolutionary search process more efficiently, a proxy dataset is used to evaluate the fitness of candidate solutions, rather than using the full dataset for training. Two popular strategies are used to create the proxy dataset. The first method involves selecting only a small proportion of the training dataset. A uniform distribution is used to select the most representative subset of the original dataset, which reduces the size of the fitness evaluation dataset while ensuring that the sample subset retains the key characteristics of the entire training dataset, potentially reducing the cost of evaluating the performance of candidate architectures. Another approach is to reduce the size of the original images, which is known as downscaling images. Such downscaled image datasets, from high-resolution to lower resolutions, as in benchmark datasets like CIFAR-10, can be easily visualized by humans and accurately replace the actual dataset. The downscaling strategy can also be used to evaluate candidate solutions during the evolutionary process \cite{kim2018task}.

The proposed methodology employs the downscaled proxy dataset to reduce the time needed for expensive fitness evaluations during the search process. Therefore, another hyperparameter, the downscaling factor \textit{n}, is required. The ratio of image size reduction is known as the downscaling factor. For instance, an image is downscaled from \textit{w × h} to \textit{w/n × h/n}. Training the dense block for one epoch reduces the feature map size to \textit{1/n × 1/n} of the original by downscaling the image, thereby reducing the memory and processing time needed to train the block to \textit{1/n × 1/n} of the actual size. 

A sensitivity analysis regarding the downscaling factor $n$ was conducted (see Appendix \ref{app:downscaling_sensitivity}, Table~\ref{tab:proxy-sensitivity}). The findings indicate that altering $n$ between 2 and 8 shows little impact on proxy accuracy, hence validating the reliability of the selected value $n = 4$.

\subsection{Evolutionary Process}\label{sec3:Evolutionary Process}
The surrogate performance predictor, in integration with proxy datasets discussed in Sections \ref{sec3:Surrogate Performance Predictor} and \ref{sec3:Proxy Data Set}, respectively, is employed by the PSO to search for the optimum dense block. The complete evolutionary neural architecture search process is outlined in Algorithm \ref{alg:evolving_dense_block}. The existing PSO procedure may be used since the suggested encoding approach converts the variable block length parameter into a fixed-length vector to represent the dense block architecture. However, to understand the notion of the surrogate predictor in EC methods, a few points need to be clarified, especially those related to the proposed model-free surrogate predictor.

\begin{algorithm}[H]
\caption{Evolve Dense Block with MFSPNet Strategy}
\label{alg:evolving_dense_block}
\begin{algorithmic}[1]
\State \textbf{Input:} Number of generations $g$, validation-cutoff epochs $c$
\State Initialize population $\mathcal{P} = \{\boldsymbol{\xi}_1, \boldsymbol{\xi}_2, \dots, \boldsymbol{\xi}_N\}$ using Algorithm \ref{alg:population_initialization}
\State Initialize personal bests $\boldsymbol{p}_{best,i} \gets \boldsymbol{\xi}_i$, $\forall i$
\State Initialize global best $\boldsymbol{g}_{best} \gets \arg\min_{\boldsymbol{\xi}_i \in \mathcal{P}} f(\boldsymbol{\xi}_i)$
\State $j \gets 0$
\While{$j < g$}
    \For{each particle $\boldsymbol{\xi}_i \in \mathcal{P}$}
        \State Update velocity $\boldsymbol{\upsilon}_i$ and position $\boldsymbol{\xi}_i$ using Eqs.~\ref{eq_velocity} and \ref{eq_position}
        
        \State Evaluate fitness:
        \Statex \hspace{1.5em} $f(\boldsymbol{\xi}_i) \gets$ VLE-EMA performance after training for $c$ epochs (Eq.~\ref{performance_estimator})
        
        \If{$f(\boldsymbol{\xi}_i) < f(\boldsymbol{p}_{best,i})$}
            \State $\boldsymbol{p}_{best,i} \gets \boldsymbol{\xi}_i$
        \EndIf
    \EndFor
    
    \State Update global best:
    \Statex \hspace{1.5em} $\boldsymbol{g}_{best} \gets \arg\min_{\boldsymbol{p}_{best,i}} f(\boldsymbol{p}_{best,i})$
    
    \State $j \gets j + 1$
\EndWhile
\State \textbf{return} $\boldsymbol{g}_{best}$
\end{algorithmic}
\end{algorithm}
% \begin{algorithm}[H]
% \caption{Evolve Dense Block with  MFSPNet Strategy}
% \label{alg:evolving_dense_block}
% \begin{algorithmic}[1]
% \State \textbf{Input:} generations $g$, validation-cutoff epochs $c$
% \State $pop \gets$ Population initialization using Algorithm \ref{alg:population_initialization}
% \State $gbest, j \gets$ Empty, 0
% \While{$j < g$}
%     \For{particle $i$ in $pop$}
%         \State $i \gets$ Update particle $i$ using PSO operations (Eqs. \ref{eq_velocity} and \ref{eq_position})
%         \State $fitness \gets$ Train $i$ for $c$ epochs and predict the performance \Statex \hspace{8em} using Eq. \ref{performance_estimator}
%         \If{$fitness_i < fitness_i(\boldsymbol{p}_{best})$}
%             \State Update particle $\boldsymbol{\xi}_i$ to $\boldsymbol{\xi}_i^{new}$  
%             \State Update $fitness_i(\boldsymbol{p}_{best})$ to $fitness_i$
%         \EndIf
%     \EndFor
%     \State $gbest \gets$ Update $\boldsymbol{g}_{best}$ and $pop$ for generation $j$ 
%     \State $j \gets j + 1$
% \EndWhile
% \State \textbf{return} $\boldsymbol{g}_{best}$
% \end{algorithmic}
% \end{algorithm}

\begin{enumerate}
    \item The surrogate predictor is model-free and does not need to be trained during the search, as described in Section \ref{sec3:Surrogate Performance Predictor}. It uses the validation losses to estimate whether the updated particle position, representing a candidate architecture, will perform better than its own optimal position, i.e., the personal best ($\boldsymbol{p}_{best}$). The candidate solution is updated with a new solution only if the new solution outperforms the personal best ($\boldsymbol{p}_{best}$).
    \item The \textit{cutoff epochs}, which is the same as the one referred to in Section \ref{sec3:Population Initialization}. 
\end{enumerate}
Consequently, the suggested approach saves computational costs by alleviating the need for full training of candidate solutions during each generation.

\subsection{Block Stacking and Domain Transfer}\label{sec3:Block Stacking}
Although block-based search reduces the search space, naively stacking multiple blocks can lead to excessive computational cost and gradient instability due to repeated channel concatenation. Deep stacking may also weaken feature reuse across distant layers. Therefore, an efficient stacking mechanism is required that allows depth scaling without linearly increasing FLOPs or parameters. To address this, we propose a decayed weighted residual aggregation strategy that promotes long-range feature reuse while maintaining nearly constant computational complexity. Since the dense block acquired with the evolutionary approach outlined in Algorithm \ref{alg:evolving_dense_block} may not be optimal to accurately reflect the complexities of the entire dataset, as it is learned from a proxy dataset, a block stacking strategy, shown in Algorithm \ref{alg:stacking_evolved_blocks}, is presented to improve the learning capacity of the final network architecture. Moreover, the capabilities of the dense block learned on one dataset might not generalize to other datasets; hence, the stacking strategy is also necessary for domain transfer. The stacking strategy, based on the block-based skip-connection, is employed to improve the learnability of the block for different dataset domains \cite{benmeziane2023skip}. This block-based skip-connection strategy is represented by Eq. \ref{skipConBlock}.

\begin{equation}\label{skipConBlock}
\mathbf{H}_l=\mathbf{F}_l\left(\mathbf{H}_{l-1}\right)+\sum_{i=1}^{l-1} \alpha_i {P}_i (\mathbf{H}_i)
\end{equation}
Where $\mathbf{H}_l$ represents the output of the $l_{th}$ dense block by aggregating previous block outputs through a weighted residual summation from the $(l-1)_{th}$ and all preceding blocks. Since the outputs of different blocks may have varying channel dimensions, each $\mathbf{H}_i$ is passed through a $1\times1$ projection layer, represented by $P_i$, to match the dimensionality of $\mathbf{H}_{l-1}$ prior to weighted summation. This ensures shape consistency for element-wise aggregation. $\mathbf{F}_l$ denotes the standard dense block operation applied on the $(l-1)_{th}$ block. The $\alpha_i$ represents the weight factor applied to the aggregation of all preceding dense blocks, ensuring stronger influence from recently stacked (nearer) blocks while progressively reducing the contribution of earlier (farther) blocks. The weight factor $\alpha_i$ can be represented with Eq. \ref{alpha}.  

\begin{equation}\label{alpha}
\alpha_i=\frac{e^{-\psi(l-i)}}{\sum_{k=1}^{l-1} e^{-\psi(l-k)}}
\end{equation}
where the weight decaying factor $\psi$ adjusts the influence of preceding blocks. Since $\psi > 0$, the exponential term $e^{-\psi(l-i)}$ decreases as the distance $(l-i)$ increases. Therefore, blocks that are closer to the current stacking position contribute more strongly, while earlier blocks have progressively smaller influence. The normalization term ensures that all aggregation weights sum to one, maintaining stable signal scaling across depths. 

The block-based dense connection can be mathematically expressed with Eq. \ref{skipConBlock2}.  
\begin{equation}\label{skipConBlock2}
\mathbf{H}_l = \mathbf{F}_l(\mathbf{H}_{l-1}) + \sum_{i=1}^{l-1} \left( \frac{e^{-\psi (l-i)}}{\sum_{k=1}^{l-1} e^{-\psi (l-k)}} \right) {P}_i (\mathbf{H}_i)
\end{equation}

The stacking process employs a weighted residual aggregation mechanism, where the output of each dense block is projected and adaptively fused with the outputs of all preceding blocks. This enables effective information flow across depth, promotes adaptive feature reuse while preventing early-block dominance, thereby stabilizing gradient propagation and improving optimization in deeper stacked architectures \cite{benmeziane2023skip}. The final CNN architecture may comprise a variable number of dense blocks to efficiently learn complex patterns. The \textit{timestostack} ($T_{max}$) of the evolved block is another hyperparameter that is adjusted according to the dataset's complexity to limit the depth of the final architecture.  

To accelerate the stacking procedure, a set of candidate architectures is produced by stacking the evolved block up to $T_{max}$ times. These candidates are eventually transferred to GPU cards for evaluation. During this training process, each candidate architecture is evaluated on the entire training dataset, which is divided into training and test sets. The training set is used to train the candidate architectures, and the evaluation is performed on the test set. The optimal architecture is chosen based on the classification accuracies acquired for all architectures during the stacking process.

%%%%%%%%% ALGORITHM BLOCK STACKING %%%%%%%%%%%
\begin{algorithm}[H]
\caption{Block Stacking with Decaying Weighted Residual Aggregation}
\label{alg:stacking_evolved_blocks}
\begin{algorithmic}[1]
\State \textbf{Input:} PSO–evolved block $b$, max stack depth $T_{max}$, dataset $D$
\State \textbf{Output:} Best stacked model $S_{best}$
\State $S_{set} \gets \emptyset$
\State Split $D$ into $D_{train}$ (80\%) and $D_{test}$ (20\%)

\For{$t = 1$ to $T_{max}$}
    \State Initialize stack $S_{current} \gets \emptyset$
    \State $H_1 \gets b$; append $(b, H_1)$ to $S_{current}$
    \For{$k = 2$ to $t$}
        \State Project previous outputs: $\tilde{H}_i = W_{proj}(H_i), \; i < k$
        \State Weighted residual aggregation:
        \[
        H_k = b + \sum_{i=1}^{k-1} \alpha^{\,k-i} \tilde{H}_i
        \]
        \State Append $(b, H_k)$ to $S_{current}$
    \EndFor
    \State Add $S_{current}$ to $S_{set}$
\EndFor

\State Train all stacked models in $S_{set}$ using Adaptive Adam on $D_{train}$ and evaluate on $D_{test}$
\State $S_{best} \gets$ stack with the highest validation accuracy
\State \Return $S_{best}$
\end{algorithmic}
\end{algorithm}

% NEW COMMNADS TO GENERATE TABLES %
\def\thickhline{\noalign{\hrule height.4pt}}
\renewcommand{\arraystretch}{1.1} %Cell Height Scaling Command
%%%%%%%%%%%%%%%%%%%%%%%%%%%%%%%%%%%%%%%%%%%%%%%%%

%% DESIGN OF EXPERIMENTS %%
\section{Design of Experiments}\label{sec4:Experiments}
The details of the experimental design are described in this section. The results of the proposed study are evaluated on four benchmark datasets: CIFAR-10, CIFAR-100, SVHN, and ImageNet. These benchmark datasets, along with peer competitors, will be discussed in Sections \ref{sec4:Benchmark_datasets} and \ref{sec4:Peer Comparison}. Moreover, Section \ref{sec4:Experimental Settings} will describe the experimental settings and parameter ranges used to carry out the experiments.

\subsection{Benchmark Data Sets}\label{sec4:Benchmark_datasets}
Initially, the dataset used by the proposed technique to create the proxy dataset must be obtained. CIFAR-10 is a widely used benchmark dataset for image classification. It consists of a medium-scale collection of images divided into ten classes, each with a moderate level of complexity. This makes it a suitable choice for evaluating image classification tasks \cite{krizhevsky2009learning}. 

We acknowledge the concern that utilizing CIFAR-10 as both the proxy for architectural search and as a final evaluation benchmark might induce benchmark-selection bias. To address this, the evolved block was evaluated on multiple, diverse datasets (CIFAR-100, SVHN, and ImageNet), whose results are reported in Table \ref{tab:TblComparisonTOP_K}. The consistent performance across various datasets demonstrates that the proxy-guided search provides designs that transfer beyond CIFAR-10. However, this constraint is acknowledged, and the transferability is empirically confirmed on larger and more diversified benchmarks. Using a lightweight proxy like CIFAR-10 is a viable practical choice given the computational cost of NAS-style surrogate modeling \cite{wang2021surrogate}.

Consequently, there are three advantages to preparing the proxy dataset with CIFAR-10: 
\begin{itemize}
    \item The first advantage is that the variety of image instances can be reflected in the proxy from the CIFAR-10 dataset.
    \item Secondly, low computing resources are required for training CNNs on the proxy dataset.
    \item Lastly, the small number of classes in the CIFAR-10 dataset speeds up the fitness evaluation process.
\end{itemize}

The CIFAR-10 dataset includes 60,000 RGB images comprised of 10 classes, with 50,000 training images and 10,000 test images. Furthermore, CIFAR-10 is employed once again to evaluate the block's classification performance and confirm that the evolved block, which was learned from the downscaled version of itself, is effective. Finally, the evolved block will also be evaluated with other benchmark datasets for transferability assessment. CIFAR-100 \cite{krizhevsky2009learning} was chosen for evaluating transferability due to its increased difficulty, as it has 100 classes compared to the smaller number of classes in other datasets, making the classification task more challenging. In addition to CIFAR-100, the SVHN dataset \cite{yuval2011reading}, which consists of digit images from 0 to 9 extracted from house numbers in Google Street View images, will be used to evaluate the transferability of the evolved block in a different domain. For further assessment, the ImageNet dataset \cite{KrizhevskyUnknown}, containing 1.2 million images across 1000 classes, will be utilized to test the transferability of the proposed approach on a much larger scale.

\subsection{Peer Comparison}\label{sec4:Peer Comparison}
The selection of comparable competitors is based on two key factors: the availability of performance metrics on benchmark datasets and their relevance to the proposed methodology. To evaluate the effectiveness of the evolved block, we first assess its performance on the CIFAR-10 dataset. Additionally, we investigate whether evolving the block on its proxy dataset impacts its performance. For a fair comparison, the competing methods are categorized into three groups. 

The first group includes two of the most sophisticated CNNs that have been crafted manually: ResNet \cite{He2015} and DenseNet \cite{HuangUnknown}. The CNNs in the second group are built automatically by employing RL methods: PNASNet \cite{liu2018progressive}, BlockQNN \cite{ZhongUnknown}, EAS \cite{Cai2017}, NASNet-A (7 @ 2304) \cite{Zoph2017}, NASH (ensemble across runs) \cite{elsken2017simple}, and NAS v3 max pooling \cite{zoph2016neural}. The last group of peer competitors is designed automatically with different EC techniques, which include EIGEN \cite{ren2019eigen}, RENAS \cite{Chen2018}, AECNN \cite{8742788}, AmoebaNet-B (6,128) \cite{Real2018}, Hier. repr-n, evolution (7000 samples), CGP-CNN (ResSet) \cite{Suganuma_2017}, DENSER \cite{Assun_o_2018}, GeNet from WRN \cite{Xie2017}, CoDeepNEAT \cite{Miikkulainen_2019}, LS-Evolution \cite{real2017large}, and EffPNet \cite{wang2021surrogate}. 

The relevance of the proposed EC approach with peer competitors increases from the first to the last group, which explains the larger number of competitors in the last group. It is also necessary to confirm how well the evolved block performs when transferred to different datasets. Therefore, based on available performance metrics on selected benchmark datasets, the following peer competitors are considered more relevant for comparison with the proposed technique: WideResNet \cite{Zagoruyko2016}, ResNet \cite{He2015}, DenseNet ($k = 12$) \cite{HuangUnknown}, CiCNet \cite{7879808}, Deeply Supervised Net \cite{Lee2014}, FractalNet \cite{Larsson2016}, Network in Network \cite{7879808}, and EffPNet \cite{wang2021surrogate}.

\subsection{Experimental Settings}\label{sec4:Experimental Settings}
The selection of hyperparameters in Table \ref{tab:TblParameters} follows three complementary strategies. 
First, the PSO parameters ($\omega$, $c_1$, $c_2$) are chosen based on commonly adopted settings in prior studies \cite{shi1998parameter, schutte2005particle, van2006study}, ensuring stable convergence behavior. 
Second, the surrogate-related parameters, such as the decay factor ($\gamma$) and validation cut-off epochs, are determined empirically through sensitivity analysis (see Section \ref{sec5:Performance with Different Epochs}), where different configurations were evaluated to balance computational efficiency and ranking reliability. Finally, architecture-specific parameters, including growth rate range, maximum layers, stacking depth ($T_{max}$), and skip-weight decay ($\psi$), are selected based on practical hardware constraints and preliminary experiments to avoid memory overflow while maintaining sufficient model capacity. 
This combination ensures that the chosen parameters are both computationally feasible and empirically effective.

The complete list of hyperparameters used in the experiments is provided in Table \ref{tab:TblParameters}. The depth (i.e., number of layers) of the dense block is initially set to 16. Taking into account hardware limitations, the growth rates for the block to be evolved with the EC technique are constrained between 12 and 32. The experiments of the proposed methodology were conducted on a distributed system specifically intended find deep CNN architectures \cite{wang2019evolving}, employing multiple Nvidia Tesla P40 GPUs, which are the least expensive GPU cards used. In our initial trial studies, it was observed that runtime errors due to insufficient memory occurred more frequently if the aforementioned hyperparameters exceeded the specified ranges. Although the provided parametric ranges are not intended to be the best possible values, they were intentionally chosen based on the available computing resources and can demonstrate the efficacy of the suggested approach. If a high-power compute system is accessible, these ranges can be extended to better optimize performance. Furthermore, a selection of hyperparameters is determined for both the model-free surrogate predictor and the proxy dataset, as listed in Table \ref{tab:TblParameters}.

%%%%%%%%%%% PARAMETERIC TABLE %%%%%%%%%%%%%%%%%%%%%%%%%
\begin{table}[b]
\centering
\caption{Experiment parameters and their corresponding values.}
\label{tab:TblParameters}
\begin{tabular}{p{0.65\linewidth} p{0.25\linewidth}}
\hline
\textbf{Parameter Type} & \textbf{Value} \\
\hline
\multicolumn{2}{l}{\textbf{PSO Hyperparameters}} \\
\hspace{3mm}Weight-inertia ($\omega$) & 0.83 \\
\hspace{3mm}Cognitive weight ($c_1$) & 1.17 \\
\hspace{3mm}Social weight ($c_2$) & 1.69 \\
\hline
\multicolumn{2}{l}{\textbf{Surrogate Performance Estimator}} \\
\hspace{3mm}Loss decaying factor ($\gamma$) & 0.78 \\
\hspace{3mm}Validation cut-off epoch & 15 \\
\hline
\multicolumn{2}{l}{\textbf{MFSPNet Hyperparameters}} \\
\hspace{3mm}Max. layers in dense block (Sec.~\ref{sec3:Particle Encoding}) & 16 \\
\hspace{3mm}Growth rate range (Sec.~\ref{sec3:Particle Encoding}) & [12–32] \\
\hspace{3mm}Dataset downscaling factor (Sec.~\ref{sec3:Proxy Data Set}) & 4 \\
\hspace{3mm}Max. stacking evolved block (Sec.~\ref{sec3:Block Stacking}) & 8 \\
\hspace{3mm}Skipping weight decay $(\psi)$ (Sec.~\ref{sec3:Block Stacking}) & 0.88 \\
\hline
\end{tabular}
\end{table}

A threshold of \textit{cutoffepochs} for fitness evaluation of candidate solutions using the surrogate predictor is set to 15. To ensure diversity during the evolutionary process, the cutoff epoch is adjusted based on previous experiments, as discussed in Section \ref{sec5:Performance with Different Epochs}. The downscaling factor for the proxy dataset (i.e., small-sized and low-dimensional dataset) is set to 2. The \textit{timestostack} ($T_{max}$) of the evolved block and the \textit{decaying-factor} ($\psi$) of the block-based skip-connection during stacking are set to 8 and 0.88, respectively. Both of these stacking parameters can be adjusted according to the complexity of the dataset. Finally, the PSO parameters are configured to generally accepted levels based on standards within the research community \cite{shi1998parameter, schutte2005particle, van2006study}.

\subsection{Compute Budget and Search Cost Accounting}
\label{sec:compute_budget}

To ensure a fair and reproducible evaluation, we provide a strict compute-budget accounting for MFSPNet. All search experiments were conducted on a single \textbf{NVIDIA Tesla P40 (24 GB)} GPU using FP32 precision, without mixed-precision (AMP), distillation, or advanced data augmentations. Candidate architectures were trained using a proxy resolution of 8$\times$8 with a batch size of 256 and the Adam optimizer at a fixed learning rate of 0.001. Each candidate was trained for 15 epochs, corresponding to approximately 195 training steps per candidate. The PSO search employed a population of 20 particles over 55 generations, resulting in a total of 1100 fully evaluated candidates. Each search run required approximately 3.5 hours, corresponding to a total of ~1.46 GPU-days over ten independent runs. Table~\ref{tab:compute_budget} summarizes the complete and reproducible compute-budget for MFSPNet, providing all relevant details required for strict performance and cost comparisons.

\begin{table}[t]
\centering
\caption{Strict compute-budget accounting for MFSPNet search (proxy evaluation).}
\label{tab:compute_budget}
\begin{tabular}{l c}
\hline
\textbf{Specification} & \textbf{Value} \\
\hline
GPU model & NVIDIA Tesla P40 (24 GB) \\
Batch size & 256 \\
Resolution (search) & 8$\times$8 (proxy) \\
Training epochs per candidate & 15 \\
Training steps per candidate & $\approx$195 \\
PSO population size & 20 \\
Number of generations & 55 \\
Fully evaluated candidates & 1100 \\
Wall-clock time per run & $\approx$3.5 hours \\
Total for 10 runs & $\approx$35 hours = 1.46 GPU-days \\
Precision & FP32 (no AMP) \\
Data augmentation & Standard CIFAR (no RA/AA) \\
Distillation & Not used \\
\hline
\end{tabular}
\end{table}

This reporting ensures transparency and allows strict, reproducible comparisons with NAS methods that use different hardware, training pipelines, or augmentations. MFSPNet completes the search in under 5.2 GPU-days while maintaining competitive performance.

\section{Results and Analysis}\label{sec5:Results}
This section presents an analysis of the experimental results. Initially, the error rate of the four benchmark datasets—CIFAR-10, CIFAR-100, SVHN, and ImageNet—will be examined and compared with that of peer competitors for a comprehensive result analysis. Subsequently, the convergence of the proposed methodology will be examined and represented graphically. Furthermore to investigate the efficacy of the proposed surrogate predictor, the analysis of fitness evaluation during evolutionary generations will be conducted. Lastly, the effect of different growth rates at each layer during PSO generations will be visualized to determine the variations of growth rates at each layer of the evolved block.

\subsection{Performance Analysis}\label{sec5:Performance Analysis}
\subsubsection{Block Assessment on CIFAR10 Dataset}\label{sec5:Transferability_Assessment}
Table \ref{tab:TblComparisonCIFAR_10} lists the classification error rate, total trainable parameters, and the computational time needed to achieve the final CNN using the proposed approach, alongside the selected peer competitors. 
For the error rate percentage and number of trainable parameters columns, if the result is obtained from multiple experimental runs, the mean value and standard deviation will be reported as mean \textpm \textit{standard deviation}.  In terms of classification accuracy, the proposed method ranks fourth among the 19 competing peers. The difference between MFSPNet and the top three methods is marginal, indicating that competitive accuracy is achieved with a substantially smaller search budget. Under the same GPU-day constraint (less than 3 GPU-days), MFSPNet achieves a comparable mean error with reduced variance across 10 independent runs.
The Mann–Whitney–Wilcoxon (MWW) test comparing MFSPNet (10 runs) with DenseNet \cite{HuangUnknown} (reported error rate 3.74\%) indicates statistical significance at the 5\% level, confirming that the observed performance difference is unlikely due to random variation. Regarding the number of parameters, the smallest CNN discovered by the proposed technique had a slightly higher number of parameters compared to the fourth-ranked model among the nineteen competing models. Nevertheless, the two competitors, despite their comparable sizes, exhibit much worse computational cost as compared to the suggested approach.

Moreover, several other approaches, such as NASNet-A, PNASNet, BlockQNN, EAS, and NAS v3—represent some of the earliest and most influential RL based strategies in automated architecture design. As illustrated in Table \ref{tab:TblComparisonCIFAR_10}, RL-based methods often achieve competitive accuracy (e.g., NASNet-A: 2.97\%, PNASNet: 3.41\%), at the expense of extremely large computational budgets. For instance, NASNet-A requires approximately 2,000 GPU-days, NAS v3 consumes 22,400 GPU-days, and BlockQNN requires 96 GPU-days for search. In contrast, the proposed MFSPNet attains an error rate of 3.76\% while consuming fewer than 3 GPU-days, with remarkable computational budget. Although some RL-based methods yield marginally lower error rates, their prohibitive search cost makes them impractical for most real-world applications. Therefore, when computational cost is explicitly constrained, MFSPNet provides a favorable accuracy-to-search-cost tradeoff, achieving competitive error rates under substantially reduced GPU-day budgets compared to RL-based NAS approaches.

\begin{table}[t]
\centering
\caption{Comparison of error rate, trainable parameters, FLOPs, and computation cost with peer competitors trained on CIFAR-10.}
\label{tab:TblComparisonCIFAR_10}
\begin{tabular}{p{0.26\linewidth} p{0.14\linewidth} p{0.12\linewidth} p{0.15\linewidth} p{0.18\linewidth}}
\hline
\textbf{Peer Methods} & \textbf{Parameters (M)} & \textbf{FLOPs (G)} & \textbf{Error (\%)} & \textbf{GPU Days} \\
\hline
DenseNet (k=40) \cite{HuangUnknown}                    & 27.20 & --   & \textbf{3.74} & -- \\
ResNet-110 \cite{He2015}                               & \textbf{1.71} & --   & 6.43 & -- \\
\hline
NASNet-A (7@2304) \cite{Zoph2017}                       & 27.60 & 3.30 & \textbf{2.97} & 2000 \\
NASH (ensemble) \cite{elsken2017simple}                 & 88.00 & --   & 4.40 & 4 \\
EAS (Transformation) \cite{Cai2017}                     & 23.40 & --   & 4.23 & $<10$ \\
BlockQNN \cite{ZhongUnknown}                            & 39.80 & --   & \textbf{3.54} & 96 \\
PNASNet \cite{liu2018progressive}                       & 3.20  & 3.20 & \textbf{3.41}$\pm$0.09 & 225 \\
NAS v3 (max pooling) \cite{Zoph2017}                    & 7.10  & --   & 4.47 & 22400 \\
\hline
EIGEN \cite{ren2019eigen}                               & \textbf{2.60} & --   & 4.40 & \textbf{2} \\
Hierarchical Rep. Evolution \cite{liu2017hierarchical}  & --    & --   & \textbf{3.75} & 300 \\
RENAS \cite{Chen2018}                                   & 3.50  & --   & \textbf{2.88} & 6 \\
AmoebaNet-B \cite{Real2018}                             & 34.90 & 3.20 & \textbf{2.98} & 3150 \\
CGP-CNN \cite{Suganuma_2017}                            & \textbf{1.68} & --   & 5.98 & $\sim$29.5 \\
AECNN \cite{sun2019completely}                          & 2.00  & --   & 4.30 & 27 \\
GeNet (WRN) \cite{Xie2017}                              & --    & --   & 5.39 & 100 \\
CoDeepNEAT \cite{Miikkulainen_2019}                     & --    & --   & 7.30 & -- \\
DENSER \cite{Assun_o_2018}                              & 10.81 & --   & 5.87 & -- \\
LS-Evolution \cite{Real2018}                            & 40.40 & --   & 4.40 & $>2730$ \\
EffPNet (best acc.) \cite{wang2021surrogate}            & 2.54  & --   & \textbf{3.49} & $<3$ \\
\hline
\textbf{MFSPNet (best acc.)}                            & 2.63  & 3.27 & 3.76 & $<3$ \\
\textbf{MFSPNet (10-run avg.)}                          & 2.72$\pm$0.15 & --   & 3.91$\pm$0.03$^*$ & $<3.2\pm0.12$ \\
\hline
\end{tabular}
\footnotetext[1]{* Error rates averaged over 10 runs. Statistical significance determined using the Mann–Whitney–Wilcoxon (MWW) test ($p<0.05$).}
\end{table}

\subsubsection{Performance on CIFAR-100, SVHN and ImageNet Dataset}\label{sec5:Block_Trasnferability_Performance_Assessment}
In order to validate the efficacy of the block’s transferability, the block evolved on CIFAR-10 proxy dataset is stacked multiple times to learn the complex images with multiple classes.The optimal CNN architecture is determined by evaluating the stacking performance on CIFAR-100, SVHN, and ImageNet datasets, by following the methodology outlined in Section \ref{sec3:Block Stacking}. Table \ref{tab:TblComparisonTOP_K} compares the \% of classification error on CIFAR-100, SVHN, and ImageNet to those of their peers. 
To evaluate whether the performance improvements of MFSPNet over baseline reported and reproduced methods are statistically significant, we applied the Mann–Whitney– Wilcoxon (MWW) test on the error rates obtained from 10 independent runs of MFSPNet and the 3 reproduced runs for each baseline (DenseNet, ResNet, EffPNet and DARTS). On CIFAR-100, the comparison between MFSPNet and DenseNet yields a $p-value$ of 0.008, indicating statistical significance at the 5\% level. The corresponding Cliff’s delta value of 0.65 reflects a medium-to-large effect size, suggesting that the improvement is both statistically reliable and practically meaningful under identical training settings. It is important to note that the statistical comparison is performed on an imbalanced sample size (10 runs for MFSPNet vs. 3 runs for the reproduced baselines), which may limit the statistical power of the Mann–Whitney–Wilcoxon test and affect the reliability of the significance assessment. Therefore, although the observed p-value ($p=0.008$) and effect size (Cliff’s $\delta=0.65$) suggest a meaningful performance difference, this result should be interpreted with caution rather than as conclusive statistical evidence.

On SVHN, MFSPNet achieves competitive performance relative to the reproduced baselines. However, the MWW test $(p = 0.18)$ does not indicate statistical significance under identical experimental conditions. Therefore, while the results demonstrate that the evolved block generalizes reasonably well, no claim of statistical superiority is made for this dataset.

It is important to note that due to the substantial computational requirements of ImageNet training, only a single run was conducted. Consequently, the reported error rate should be interpreted as an indicative demonstration of scalability and feasibility rather than statistically validated superiority. Multiple independent runs would be required to establish statistical confidence.  In comparison with the selected baselines, MFSPNet remains competitive on ImageNet in terms of classification accuracy. Consequently, the proposed method shows its capacity to compete even with a large dataset. The proposed method has shown general applicability by attaining rather competitive classification accuracy over three distinct standard benchmark datasets. Although the compared methods i.e. DenseNet and ResNet were not retrained under the same downscaled proxy dataset utilized in the MFSPNet. These results are based on full-scale datasets and standardized benchmark implementations found in the literature. Therefore, the comparison will serve as a reference baseline rather than an exact equivalent in experimental settings. The proposed EC search uses the proxy dataset exclusively to expedite search process. After the EC generations the evolved block is retrained on the whole dataset for fair performance evaluation. This technique verifies that the stated results of the search strategy are more of parametric variation settings.
Overall, the results indicate that MFSPNet demonstrates statistically supported improvements under equal-budget conditions on CIFAR datasets, while maintaining competitive cross-dataset transfer performance on SVHN and ImageNet without overstating statistical superiority.

The experimental findings show the possibility to transfer an evolved block from one dataset to another dataset for a similar task. The suggested technique may consist of two components that enhance transferability. First, the evolved block was selected as an optimal feature extractor from hundreds of candidate architectures. Secondly, the evolved block stacked many times using the stacking procedure outlined in Section \ref{sec3:Block Stacking} to develop a variant of CNNs with different model complexities. These CNNs were finally evaluated with a dataset of a specific domain.
Since only the best CNN was ultimately selected by choosing an appropriate stacking depth and model size, the stacking method might improve the transferability of the evolved block.
\begin{table}[ht]
\centering
\caption{Comparison of error rate (\%) on CIFAR-100 and SVHN datasets. 
For ImageNet, Top-1/Top-5 error rates are reported. 
Baselines above the midline are taken from the literature; those below are reproduced under our unified training recipe (3 runs).}
\label{tab:TblComparisonTOP_K}
\begin{tabular}{p{0.38\linewidth} p{0.15\linewidth} p{0.15\linewidth} p{0.25\linewidth}}
\hline
\textbf{Peer Methods} & \textbf{CIFAR-100} & \textbf{SVHN} & \textbf{ImageNet (Top-1/5)} \\
\hline
Network in Network \cite{7879808}                 & 35.68 & 2.35 & -- \\
DenseNet (k=12) \cite{huang2017densely}           & 20.20 & \textbf{1.67} & \textbf{25.02/7.71} \\
FractalNet \cite{Larsson2016}                     & 23.30 & 2.01 & \textbf{24.12/7.39} \\
ResNet \cite{He2015}                              & 27.22 & 2.01 & \textbf{25.03/7.76} \\
Wide ResNet \cite{Zagoruyko2016}                  & 22.07 & \textbf{1.85} & 30.04/10.93 \\
Deeply Supervised Net \cite{Lee2014}              & 34.57 & 1.90 & 33.70/13.10 \\
CiCNet \cite{7879808}                             & 24.82 & --   & -- \\
EffPNet (Best) \cite{wang2021surrogate}           & 18.49 & 1.82 & 27.01/9.28 \\
\hline
\multicolumn{4}{l}{\textit{Reproduced under identical training recipe (3 runs)}} \\
\hline
DenseNet \cite{huang2017densely}                  & 20.79$\pm$0.17 & 1.71$\pm$0.26 & -- \\
ResNet \cite{He2015}                              & 27.31$\pm$0.61 & 1.99$\pm$1.01 & -- \\
EffPNet \cite{wang2021surrogate}                  & 21.01$\pm$0.13 & 1.81$\pm$1.51 & -- \\
DARTS \cite{liu2019darts}                         & 19.32$\pm$1.02 & 1.75$\pm$1.33 & -- \\
\hline
\textbf{MFSPNet (Best)}                           & \textbf{17.68} & 1.91 & 28.29/12.82 \\
\textbf{MFSPNet (10 runs)}                        & 20.19$\pm$0.15$^*$ & 1.98$\pm$0.06$^*$ & -- \\
\hline
\end{tabular}
\footnotetext[1]{* Statistical significance computed using the MWW test ($p<0.05$); Cliff’s delta included in Section~\ref{sec:stats}.}
\end{table}

% \begin{table}[ht]
% \centering
% \caption{Peer comparison of error rate (\%) on CIFAR-100 and SVHN datasets. For ImageNet, top-k error rates are reported where $k = \{1, 5\}$.}
% \label{tab:TblComparisonTOP_K}
% \begin{tabular}{p{0.35\linewidth} p{0.15\linewidth} p{0.15\linewidth} p{0.25\linewidth}}
% \hline
% \textbf{Peer Methods} & \textbf{CIFAR-100} & \textbf{SVHN} & \textbf{ImageNet Top-k (1, 5)} \\
% \hline
% Network in Network \cite{7879808} & 35.68 & 2.35 & -- \\
% DenseNet \cite{huang2017densely} \textit{(k=12)} & 20.20 & \textbf{1.67} & \textbf{25.02/7.71} \\
% FractalNet \cite{Larsson2016} & 23.30 & 2.01 & \textbf{24.12/7.39} \\
% ResNet \cite{He2015} & 27.22 & 2.01 & \textbf{25.03/7.76} \\
% Wide ResNet \cite{Zagoruyko2016} \textit{(k=12)} & 22.07 & \textbf{1.85} & 30.04/\textbf{10.93} \\
% Deeply Supervised Net \cite{Lee2014} & 34.57 & \textbf{1.90} & 33.7/13.1 \\
% CiCNet \cite{7879808} & 24.82 & -- & -- \\
% EffPNet (Best) \cite{wang2021surrogate} & 18.49 & \textbf{1.82} & \textbf{27.01/9.28} \\
% \hline
% \textbf{MFSPNet (Best)} & \textbf{17.68} & 1.91 & 28.29/12.82 \\
% \textbf{MFSPNet (10 runs)} & 20.19$\pm$0.147$^*$ & 1.98$\pm$0.056$^*$ & -- \\
% \hline
% \end{tabular}
% \footnotetext[1]{* Error rates are averaged over 10 runs. Statistical significance was determined using the Mann–Whitney–Wilcoxon (MWW) test ($p <$ 0.05).}
% \end{table}

\subsection{PSO Convergence}\label{sec5:Performance Analysis}
\begin{figure}[h]%
\centering
\includegraphics[width=1.0\textwidth]{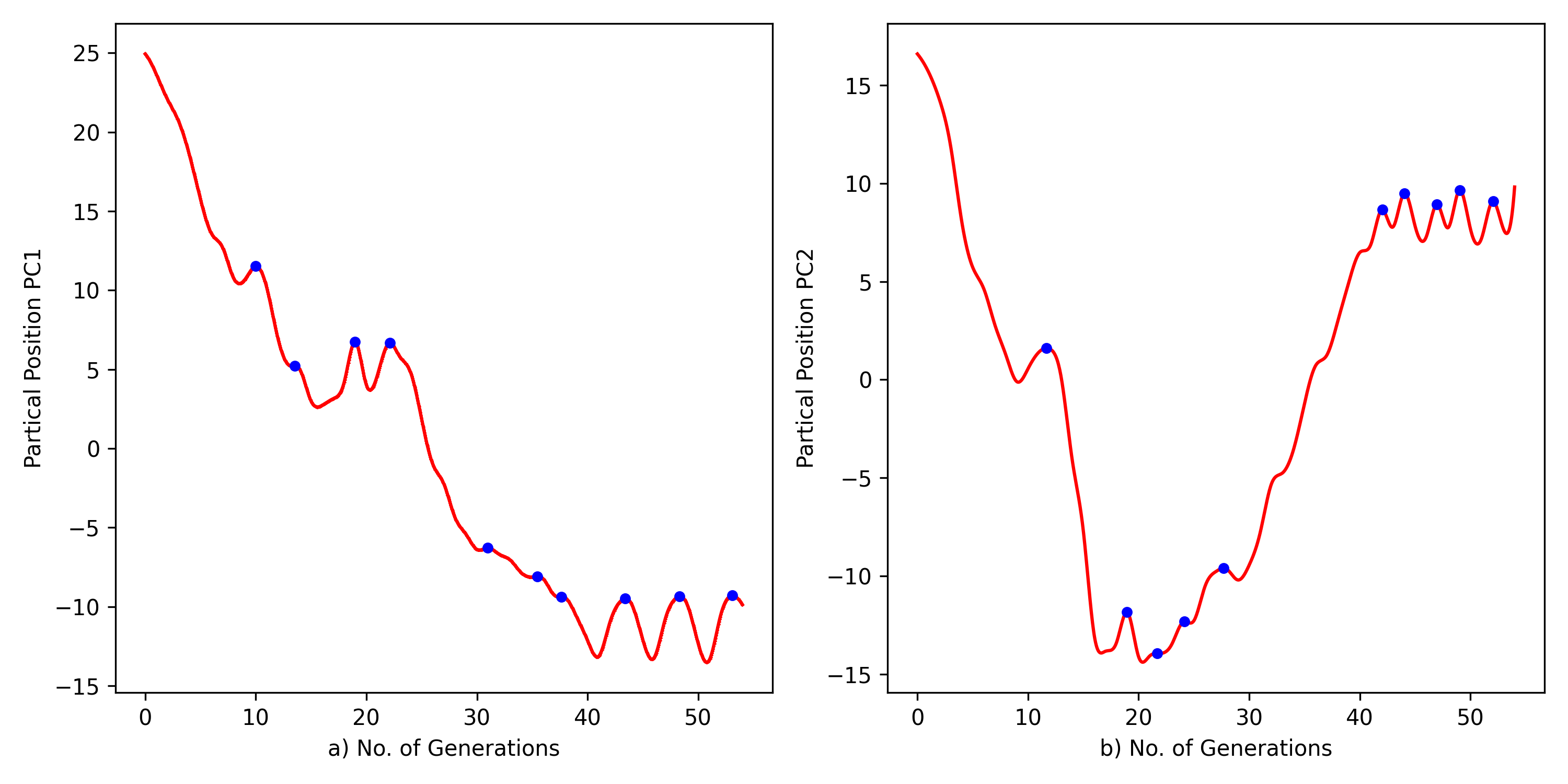}
\caption{Convergence of particle positions across generations using principal component analysis (PCA): 
(a) PC1 trajectory and (b) PC2 trajectory showing stabilization after multiple generations.}
\label{fig_dense_block}
\end{figure}

To analyze the convergence of the PSO algorithm an investigation is conducted on the positions of the particles. The principal component analysis (PCA) technique is used to represent the higher dimensional position vector i.e.(PSO particle) with 2-dimensional components \cite{wold1987principal, abdi2010principal}. Fig \ref{fig_dense_block} (a) and (b) illustrate how the particle's positions converged during evolutionary generations. It is also evident that the particle position fluctuates at the start of evolutionary generations which shows that exploration of search space. But this fluctuation becomes stable after a few generations. And at the later generations the position fluctuation curve eventually almost flattens, which reflects the transition from exploration to exploitation. The observed flattening of particle trajectories therefore indicates that the swarm collectively stabilizes around a region of high fitness, resulting  convergence to an optimal architecture configuration. The performance of the proposed surrogate-assisted PSO strategy is sensitive to parameter selections. Our study demonstrates that the inertia weight ($\omega$) affects the balance between exploration and exploitation: small values (e.g., 0.4) demonstrated the convergence but frequent trap in local minima, whereas bigger values (e.g., 0.9) enhanced exploration at a cost of slower convergence. Cognitive $(c_1)$ and social $(c_2)$ acceleration factors complement each other. The High $c_1$ values enhanced self-exploration by particles, while excessive attention prevented swarm interaction. Conversely, large $(c_2)$ values moved particles rapidly toward the global best but led to premature convergence. Finally, the population size had a direct influence on surrogate  performance: small swarms $(<20)$ lacked sufficient variety, whereas extremely large swarms $(>60)$ increased computation without improving accuracy considerably. The detailed parameter effects are summarized in ~\ref{app:pso_sensitivity}.

\subsection{Analysis of Surrogate Predictor}\label{sec5:Analysis of Surrogate Predictor}
\subsubsection{Analysis of Loss Landscape}\label{sec5:Loss_landscape}
\begin{figure}[h]%
\centering
\includegraphics[width=0.75\textwidth]{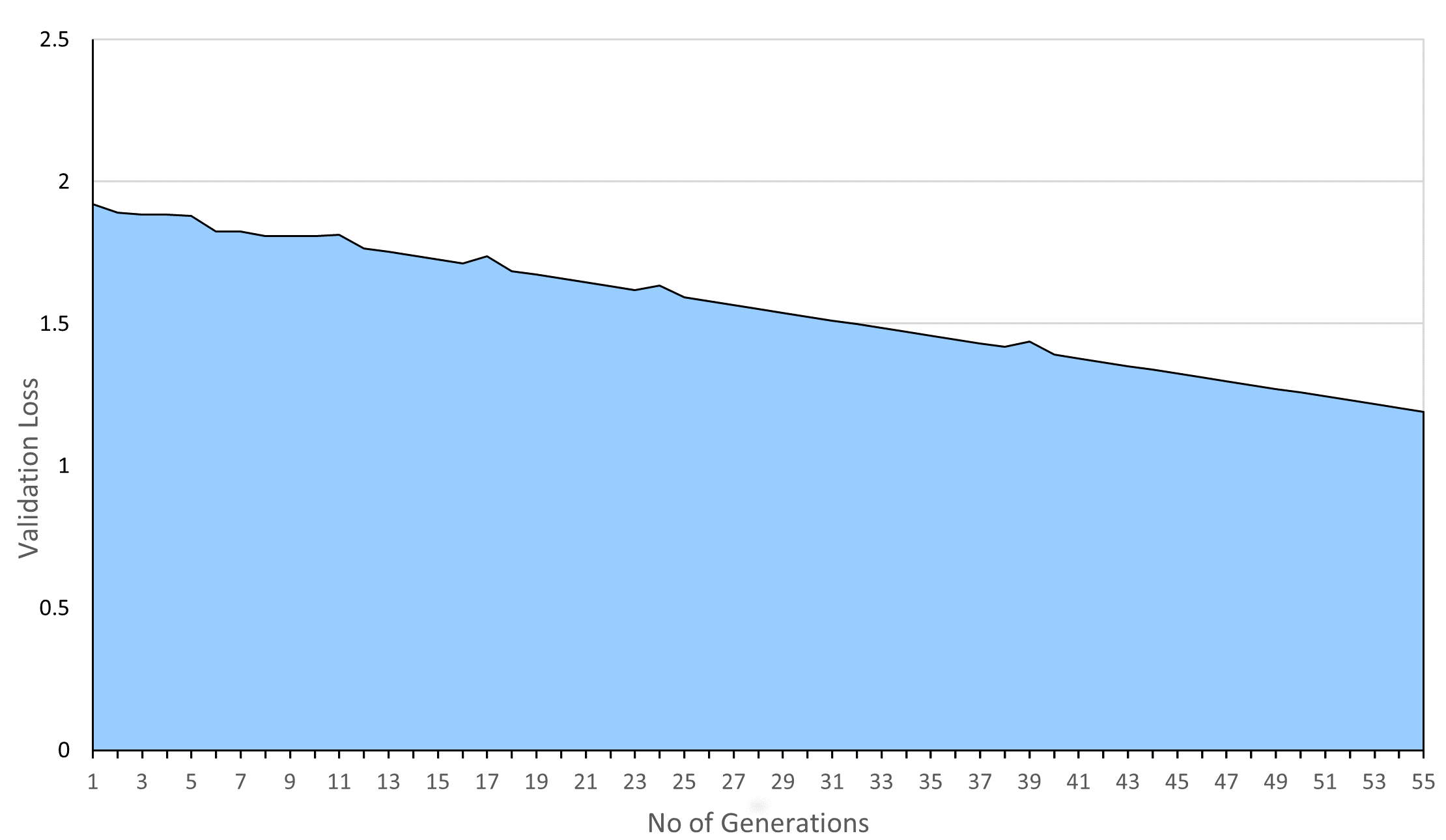}
\caption{Convergence of objective function in evolutionary generations during evolutionary generations: \textit{x-axis} and \textit{y-axis} show number of generations and validation loss of \textit{VLEEMA} }\label{Fig_Generation_Convergence}
\end{figure}

As described in Section \ref{sec2:Surrogate Performance Predictors}, the model-free surrogate performance predictor is employed for performance evaluation of candidate CNN architecture, represented by PSO particle. The validation loss of 15 epochs is used to estimate the performance of CNN. Therefore, the performance of the proposed model-free surrogate estimator can be evaluated with the trend of loss landscape during each generation of the evolutionary process. Fig. \ref{Fig_Generation_Convergence} illustrates the validation loss acquired with surrogate performance predictor across the evolutionary generations, which remarkably shows the promising performance and convergence of objective function toward global optimum position. It is evident from the loss landscape that at the start of evolutionary generations, the loss of the best particle in the whole population was quite high and gradually reduced with an increase in a number of generations. So this exhibits that the fitness value acquired with the proposed model-free surrogate predictor remarkably assists the entire evolutionary search process. 

% The area under the loss curve start decreasing after few evolutionary generations, which shows also shows the convergence of the algorithm to global optimum of fitness landscape.

\subsubsection{Performance with Different Epochs}\label{sec5:Performance with Different Epochs}
\begin{figure}[h]%
\centering
\includegraphics[width=0.65\textwidth]{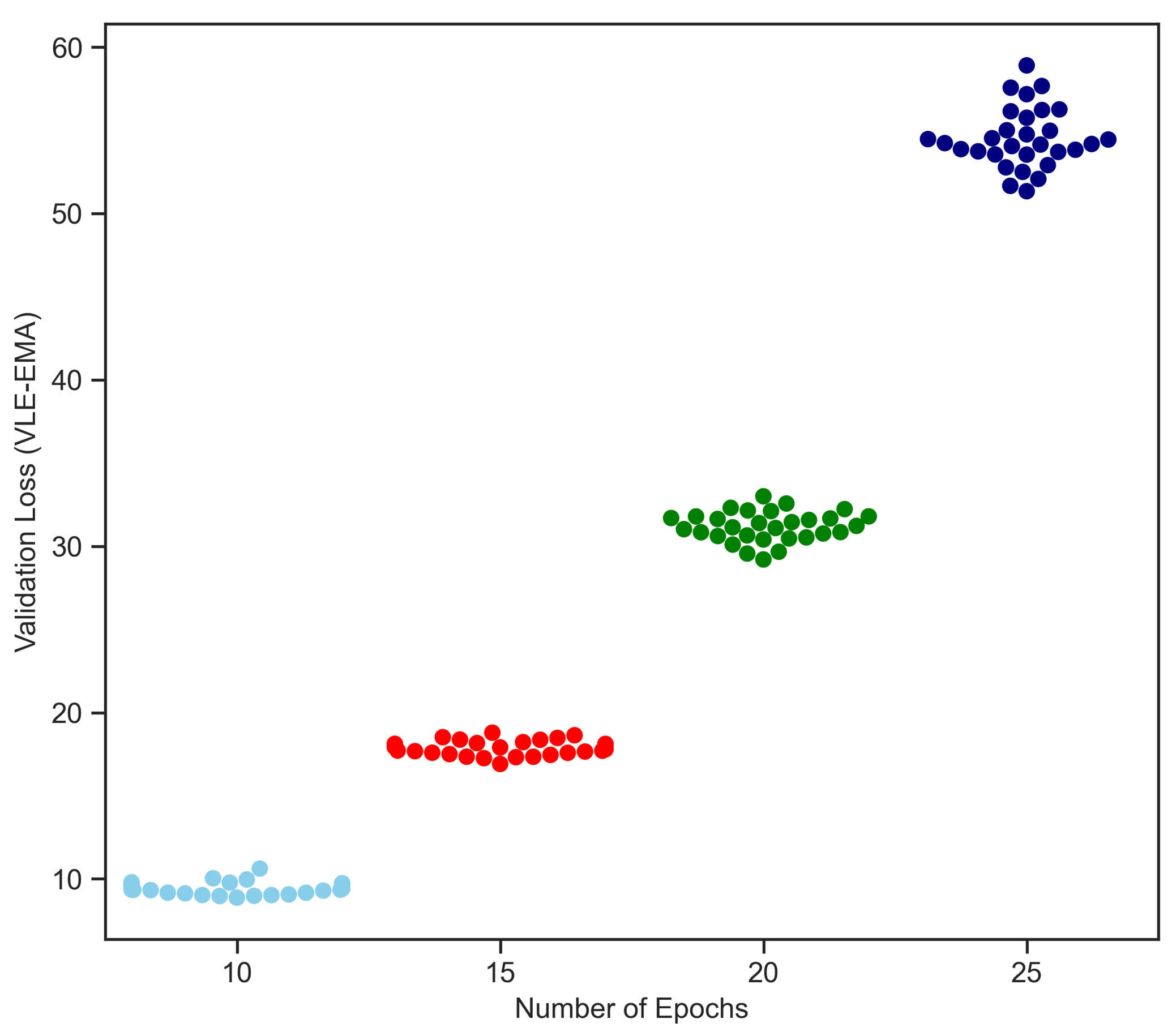}
\caption{Swarm plot to illustrate the effect of choosing different numbers of epochs to estimate the validation loss 10-25 epochs calculated using \textit{VLEEMA}}\label{Fig_Swarm_Plot}
\end{figure}

While calculating the fitness of candidate architecture the number of epochs, required to calculate fitness, plays a crucial role in the calculation of \textit{VLEEMA}. And it also influences the balance between exploration and exploitation in the PSO process. The effect of different numbers of epochs can be analyzed from several perspectives. To effectively leverage the \textit{VLEEMA} in PSO, initial experiments were performed for an effective number of epochs. It is examined during the experiments that, using fewer epochs for fitness evaluation can facilitate rapid exploration of the search space, allowing the algorithm to identify promising regions. However, at earlier epochs, the fitness estimate can mislead the search process due to insufficient network parameters and poor generalization capabilities. As the number of epochs increases, the evaluations become more stable, providing a better approximation of the network's performance and also ensuring a good balance between exploration and exploitation \cite{ru2021speedy}. Various number of epochs were evaluated for the proposed model-free surrogate performance predictor. Table \ref{tab:TblSurrogateEpochs} also shows the effect of increasing the \textit{cutoffepochs} improves the performance estimate i.e. \textit{(VLEEMA)} of CNN architecture, while with fewer \textit{cutoffepochs} the standard deviation of \textit{VLEEMA} different candidate architectures is quite low. If the \textit{cutoffepochs} is increased the diversity among the particles is increased as standard deviation of \textit{VLEEMA} will also be increased, which provides a more robust estimate of fitness. Moreover, Fig. \ref{Fig_Swarm_Plot} illustrates the swarm of fitness approximation at 10 \textit{cutoffepochs} is dilated around similar fitness value i.e.(represented at y-axis). Considering this aspect 15 \textit{cutoffepochs} provide a more robust estimate of the performance of dense block, and the swarm at 15 \textit{cutoffepochs} provides more variation in the fitness of candidate architecture.

%%%% NEW TABLE Rank Correlation between proxy %%%%
\begin{table}[t]
\centering
\caption{Rank correlation between different proxy methods and the final validation accuracy of candidate architectures. Higher correlation indicates better ranking consistency.}
\label{tab:rank_correlation}
\begin{tabular}{p{0.35\linewidth} p{0.25\linewidth} p{0.25\linewidth}}
\hline
\textbf{Proxy Method} & \textbf{Spearman $\rho$} & \textbf{Kendall $\tau$} \\
\hline
VLE-EMA (Proposed) & \textbf{0.87} & \textbf{0.71} \\
TSE-EMA & 0.63 & 0.49 \\
Low-epoch (10-epoch) training & 0.58 & 0.42 \\
\hline
\end{tabular}
\end{table}

%%%%%%%%%%% SURROGATE EPOCHS EFFECTS %%%%%%%%%%%%%%%%%%%%%%%%%
\begin{table}[ht]
\centering
\caption{Standard deviation of \textit{VLEEMA} across different training epochs.}
\label{tab:TblSurrogateEpochs}
\begin{tabular}{p{0.35\linewidth} p{0.35\linewidth}}
\hline
\textbf{Number of Epochs} & \textbf{Standard Deviation} \\
\hline
10 & 0.47 \\
15 & 0.56 \\
20 & 0.73 \\
25 & 0.79 \\
\hline
\end{tabular}
\end{table}

By recording the standard deviations among the fitness values of the initial population for varying numbers of epochs, we examine the stability and reliability of the fitness evaluations. Standard deviations of fitness values of the population for different \textit{VLEEMA} were recorded for each epoch count. Table \ref{tab:TblSurrogateEpochs} summarizes the results. The higher standard deviation indicates the variability in fitness assessment. Moreover, the lower standard deviation indicates less diversity in population fitness.
\subsection{Ranking Consistency of the Surrogate Estimator}\label{sec5:rank correlation}
To verify that the proposed VLE-EMA surrogate provides reliable guidance during evolutionary NAS, we evaluate its ability to preserve the relative performance ranking of candidate architectures. Since evolutionary search relies primarily on comparative selection rather than exact performance prediction, ranking consistency is more critical than absolute regression accuracy.
We therefore compute the Spearman’s rank correlation coefficient $(\rho)$ and Kendall’s rank correlation coefficient $(\tau)$ between surrogate-predicted performance and the final fully trained validation accuracy of sampled architectures. To ensure a fair and reproducible evaluation, the ranking analysis was conducted on a set of $N=50$ candidate architectures randomly sampled from the same search space used in the main NAS experiments. The sampled architectures include diverse block configurations with varying depths, widths, and connectivity patterns, reflecting the variability encountered during the evolutionary search process. Each architecture was first evaluated using the proposed VLE-EMA surrogate (with $c=15$ epochs), and subsequently fully trained to convergence to obtain its final validation accuracy. This consistent evaluation protocol ensures that the reported rank correlations accurately reflect the surrogate’s ability to generalize within the same search space. In addition, a rank-correlation analysis demonstrates that VLE-EMA better preserves the ordering of candidate architectures relative to final validation accuracy compared to standard proxies as illustrated in Table~\ref{tab:rank_correlation} for Spearman’s $\rho$ and Kendall’s $\tau$. Specifically, VLE-EMA achieves substantially higher Spearman ($\rho=0.87$) and Kendall ($\tau=0.71$) correlations than common proxies such as TSE-EMA ($\rho=0.63$) and low-epoch training ($\rho=0.58$), indicating that it provides a more reliable ranking of candidate architectures under identical evaluation conditions.
Therefore, strong rank consistency enables the search process to focus on promising regions of the architecture space while avoiding the heavy computational cost of full training.

\subsection{Block Stacking Analysis}\label{sec5:Block Stacking Analysis} 
The results of proposed block-based stacking illustrated in Table \ref{tab:TblComparisonTOP_K} show the efficacy of the proposed stacking approach on transferring to different dataset domains and demonstrate the evolved dense block's generalization capabilities. The term transferability does not refer to unconditional transferability to every other possible domain. The transferability across different image classification benchmarks with increased difficulty was demonstrated in Table \ref{tab:TblComparisonTOP_K}. To quantify this, a rank-preservation between proxy and target evaluations is performed and results are provided in Appendix with Table \ref{tab:proxy_correlation}. High correlation values and consistent relative rankings indicate that CIFAR-10 preserves the ordering of candidate architectures sufficiently well to guide the PSO search, while retraining on the full datasets ensures fair and comparable final performance numbers.

The generalization of the evolved block is assessed with the validation loss acquired during stacking as outlined in Algorithm \ref{alg:stacking_evolved_blocks}. The comparison with relevant peer competitors, the strategies employed the evolutionary search technique, the MFSPNet competes with them with remarkable performance. Moreover, the proposed stacking strategy is validated with multiple runs on different-scale datasets, such as CIFAR-100, SVHN, and ImageNet. These findings suggest that the evolved dense blocks can effectively capture transferable features, allowing for efficient adaptation to new domains by employing the stacking strategy outlined in Algorithm \ref{alg:stacking_evolved_blocks}.

\subsection{Effect of Growth Rates}\label{sec5:Growth Rate Analysis}
\begin{figure}[h]%
\centering
\includegraphics[width=1\textwidth]{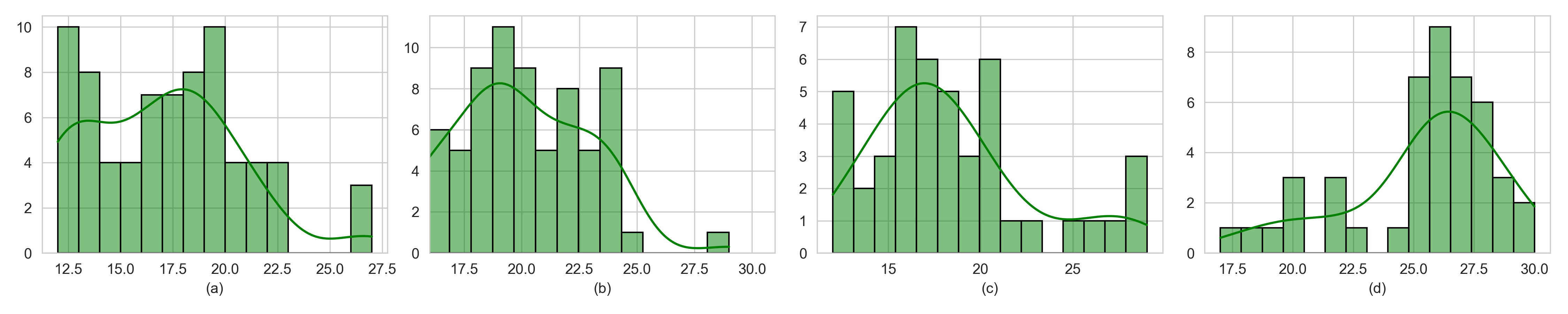}
\caption{Distribution of growth rates after all PSO generations. Growth rates distribution for initial layer, middle layers and last layer is plotted with 15 bins. \textit{x-axis} shows the growth rates and \textit{y-axis} represents the number of times the block was evaluated for a specific growth rate.}\label{fig_histogram_plot}
\end{figure}

To analyze the effects of different growth rates for different layers, the distribution plot for the evolved is illustrated in Fig.\ref{fig_histogram_plot}. By examining the distribution of growth rates for the initial, middle, and final layers, of evaluated blocks are depicted in Fig. \ref{fig_histogram_plot}(a)-(d), respectively. 

It is observed that the low growth rate bar is longer than other ranges. Conversely, the distribution for middle layers is depicted in Fig.\ref{fig_histogram_plot} (b) and (c), the mid-range growth rates are significantly more prevalent than the others. Plot (b) shows a broader distribution with a significant peak at a growth rate 20, indicating a more consistent growth rate in this layer compared to the initial layer. Plot (c), on the other hand, shows a slightly left-skewed distribution with the highest frequency captured at  18, suggesting a slight decrease in growth rates. Plot (d) represents the distribution of growth rates at the final layer, with a peak at 20 and a tail extending towards higher values.  Growth-rate distributions gradually move toward larger mean values with less variation as the network depth increase. This implies that in order to maintain feature variation and gradient flow, there is a selection pressure on designs that have broader channels in deeper layers. Theoretically, this coincides with the notion of gradient preservation in densely linked networks \cite{huang2017densely}, where larger growth rates in later layers boost reuse of features and improve convergence stability. Consequently, the broad-to-narrow development in growth-rate distribution statistically represents how the optimization process regulates architectural depth and width for balanced representation capacity.

\begin{figure}[H]%
\centering
\includegraphics[width=1\textwidth]{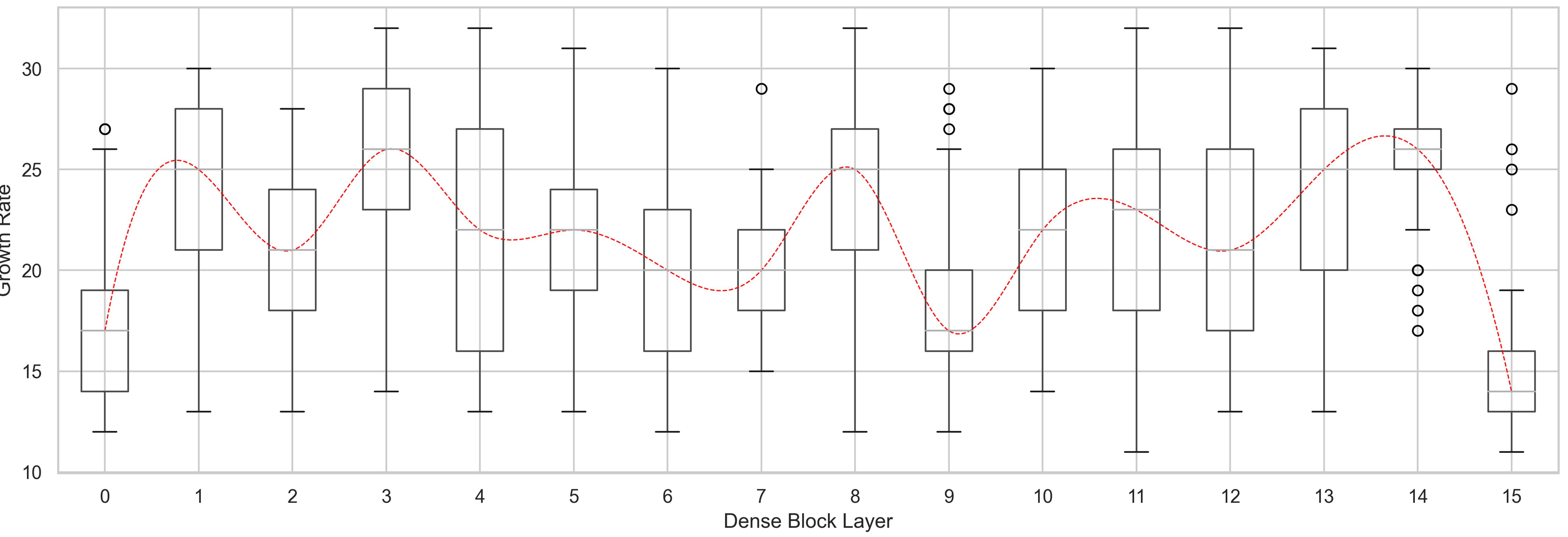}
\caption{The effect of growth rate at different layers during the evolutionary process is represented by a box plot. \textit{x-axis} represents the block layer from (1-16). And \textit{y-axis} indicates the growth rates of each layer. The median value of each layer box is depicted with a curved line.}\label{Fig_Growth_rate_analysis}
\end{figure}

Furthermore, the change in growth rates on each layer can also be seen in Fig. \ref{Fig_Growth_rate_analysis}. 
The curved line on the box plot representing each layer indicates that distinct growth rates have been selected for each layer. The first and last layers have the lowest median values, while the middle layers have comparatively high median values. 
This suggests that the intermediate layer requires a high growth rate value compared to the other, either the first or the final layer. 
Finally, the growth-rate adaption and convergence dynamics can be interpreted as combined optimization process. The PSO velocity and layer-wise growth-rate distribution ensures the stability and feature representation  capacity, respectively. The combination of these behaviors supports both efficient search convergence and the emergence of architectures that generalize well across datasets.

\noindent \textbf{Impact of Application Type and Data Content.}
Depending on the features of the dataset, the suggested surrogate-assisted PSO architecture performs differently. Rapid convergence and sustained accuracy were attained by the surrogate on CIFAR-10, which had little intra-class variability. Higher intra-class variance caused SVHN, which contains digits in noisy nature environments, to demonstrate slower convergence. Since CIFAR-100 was more fine-grained, it took more iterations to generalize well, but ImageNet's large-scale variety made it more susceptible to proxy scaling. According to these findings, surrogate accuracy and search efficiency are directly impacted by dataset content, including classification granularity, noise, and variability.

Due to the significant environmental and energy costs related to large-scale optimized deep learning model search, the proposed MFSPNet mitigate these issues by reducing the number of GPU hours required to search NAS through EC approach. As the model-free surrogate approximates the performance of model without full network training, it decreases energy consumption compared to traditional NAS approaches such as DARTS or NASNet. This efficiency improvement aligns with the growing emphasis on sustainable AI practices and resource-efficient model design.

% Finally, the proposed technique investigated the regions of the search space with the highest classification accuracy. It is important to optimize the growth rates for each layer in the dense blocks, rather than using preset growth rates for all layers, as recommended in the original DenseNet paper \cite{huang2017densely}.
\subsection{Computational Complexity Analysis}
The computational complexity of the proposed MFSPNet framework primarily arises from the PSO-based block evolution and stacked-block training stages. 
For the PSO phase, the time complexity is \(O(P \times G \times c \times N_p \times L)\), where \(P\) is the swarm size, \(G\) the number of generations, \(c\) the cutoff epochs, \(N_p\) the proxy dataset size, and \(L\) the block depth; the corresponding space complexity is \(O(P \times L)\). 
The stacking stage incurs a complexity of \(O(T_{\max} \times E \times N \times L)\), where \(T_{\max}\) denotes the number of stacked blocks, \(E\) the training epochs, and \(N\) the dataset size. 
Overall, the total computational cost scales linearly as 
\[
O(P G c N_p L + T_{\max} E N L),
\]
demonstrating efficient and scalable behavior across varying search-space sizes.

\subsection{Ablation Study}
This ablation study is designed to validate the motivations behind the two key components of MFSPNet: (i) the surrogate predictor introduced to reduce evaluation cost while preserving ranking reliability, and (ii) the decayed block stacking mechanism proposed to enable scalable depth with stable computational complexity.
Therefore, effect of major components of proposed MFSPNet, ablation experiments were performed on the CIFAR-10 proxy dataset. 

\noindent\textbf{Effect of Surrogate Predictor:} 
The proposed surrogate-assisted PSO was compared with a baseline PSO without the predictor. 
The surrogate integration reduced search time by approximately 45\% while maintaining comparable validation accuracy (within $\pm$0.3\%), indicating that the surrogate predictor effectively accelerates the search process. As described in section \ref{sec5:rank correlation}

\begin{figure}[t]
  \centering  \includegraphics[width=0.95\linewidth]{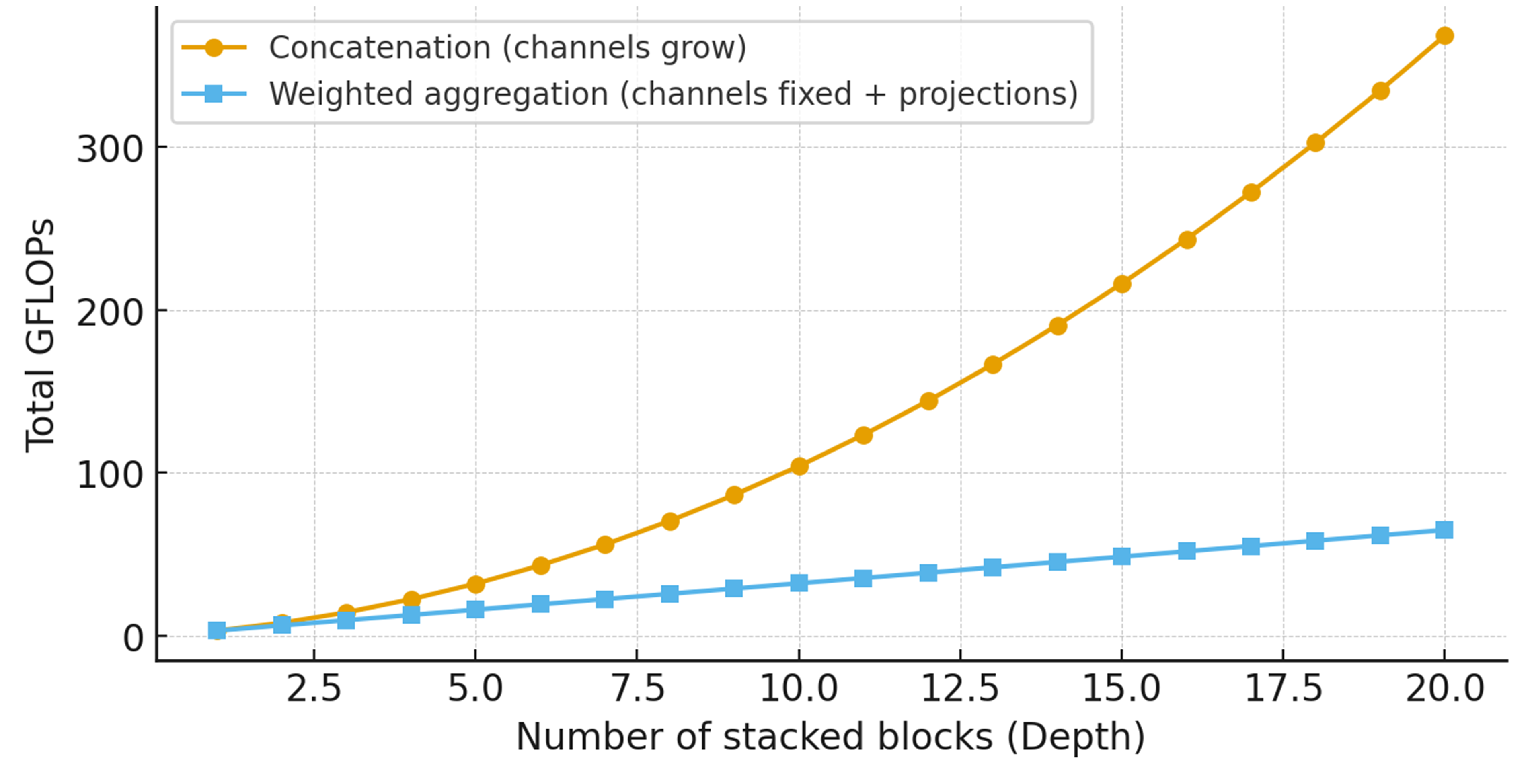}
  \caption{Total GFLOPs as a function of stacked block depth for concatenation vs. weighted residual aggregation. Weighted aggregation maintains nearly constant complexity, whereas concatenation grows linearly with depth.}
  \label{fig:flops_vs_depth}
\end{figure}

\noindent\textbf{Effect of Block Stacking:}
To evaluate the impact of the proposed stacking strategy, we conducted ablation experiments using both single and multiple stacked evolved blocks equipped with the weighted residual aggregation mechanism. The results show a consistent improvement in accuracy as the stack depth increases, confirming that the decayed multi-block aggregation promotes effective long-range feature reuse and stabilizes gradient propagation. In addition, we analyze the computational cost in terms of total GFLOPs as a function of stack depth (Fig. \ref{fig:flops_vs_depth}). Unlike channel-wise concatenation, which leads to linear growth in computational cost, the proposed aggregation maintains nearly constant complexity due to its fixed channel dimensionality and lightweight 1×1 projection. This shows that improved accuracy does not come at the expense of excessive computational overhead. Overall, these results confirm that both the surrogate predictor and the efficient stacking mechanism are essential contributors to the improved performance and search efficiency of the MFSPNet architecture.

% To evaluate the effect of block stacking, experiments with single and multiple stacked evolved blocks were performed with block-based dense connections. 
% Accuracy improved due to feature reuse and validates the effectiveness of the decaying dense connection strategy.

% \noindent Overall, these results demonstrate that both the surrogate predictor and the stacking mechanism are essential for enhancing the efficiency and performance of the MFSPNet architecture.

\section{Conclusion and Future Work}\label{sec6:Conclusion and Future Work}
This study revealed that the suggested MFSPNet framework efficiently search dense block structures with robust generalization and transferability across different datasets with medium to high level of complexity. Utilizing a downscaled proxy dataset, the method attained comparable accuracy on CIFAR-100, SVHN, and ImageNet while substantially decreasing computational cost, underscoring its efficiency and scalability in resource-limited settings. These findings validate that surrogate-assisted NAS may considerably expedite architecture discovery while maintaining performance, giving it useful for sustainable and real-time AI applications. Nonetheless, the approach presently functions inside a DenseNet-inspired search space and depends on the CIFAR-10 proxy dataset, which may not adequately reflect bigger or more diverse imagery domains. The methodology is also sensitive upon hyperparameter configurations and was assessed within certain hardware limitations that may affect scalability. Future research will concentrate on broadening the search space for hybrid CNN–Transformer architectures, improving block diversity via alternative metaheuristics, and validating the methodology across various application domains, including medical imaging, remote sensing, and other high-resolution visual tasks, to further evaluate its robustness and generalization.
\appendix
\section{Signal--Noise model and comparison of TSE-EMA and VLE-EMA}

We model the per-epoch observed loss (or metric) as a slowly-varying signal plus zero-mean noise. For validation loss at epoch $t$:
\begin{equation}\label{eq:signal_noise}
y_t^{\mathrm{val}} = \mu_t^{\mathrm{val}} + \varepsilon_t^{\mathrm{val}}, 
\qquad \mathbb{E}[\varepsilon_t^{\mathrm{val}}]=0,\quad \mathrm{Var}(\varepsilon_t^{\mathrm{val}})=\sigma_{t,\mathrm{val}}^2.
\end{equation}
An analogous decomposition holds for training loss:
\begin{equation}
y_t^{\mathrm{train}} = \mu_t^{\mathrm{train}} + \varepsilon_t^{\mathrm{train}}, 
\qquad \mathbb{E}[\varepsilon_t^{\mathrm{train}}]=0,\quad \mathrm{Var}(\varepsilon_t^{\mathrm{train}})=\sigma_{t,\mathrm{train}}^2.
\end{equation}

Both TSE-EMA and VLE-EMA are normalized weighted averages (EMA form) over $T$ observed epochs with exponential weights $w_t=\gamma^{\,T-t}$, $0<\gamma<1$.

\subsubsection*{VLE-EMA (validation-loss EMA)}
\begin{equation}\label{eq:VLE}
\widehat{S}_{\mathrm{VLE}} \;=\; 
\frac{\sum_{t=1}^T \gamma^{\,T-t}\, y_t^{\mathrm{val}}}{\sum_{t=1}^T \gamma^{\,T-t}}.
\end{equation}
Using \eqref{eq:signal_noise} the expectation and variance are
\begin{align}
\mathbb{E}\!\left[\widehat{S}_{\mathrm{VLE}}\right] 
&= \frac{\sum_{t=1}^T \gamma^{\,T-t}\, \mu_t^{\mathrm{val}}}{\sum_{t=1}^T \gamma^{\,T-t}}, \label{eq:EVLE} \\
\mathrm{Var}\!\left(\widehat{S}_{\mathrm{VLE}}\right)
&= \frac{\sum_{t=1}^T \gamma^{2(T-t)}\, \sigma_{t,\mathrm{val}}^2}
{\big(\sum_{t=1}^T \gamma^{\,T-t}\big)^2}. \label{eq:VarVLE}
\end{align}

\subsubsection*{TSE-EMA (training-loss EMA)}
\begin{equation}\label{eq:TSE}
\widehat{S}_{\mathrm{TSE}} \;=\; 
\frac{\sum_{t=1}^T \gamma^{\,T-t}\, y_t^{\mathrm{train}}}{\sum_{t=1}^T \gamma^{\,T-t}}.
\end{equation}
Analogously, its expectation and variance are
\begin{align}
\mathbb{E}\!\left[\widehat{S}_{\mathrm{TSE}}\right] 
&= \frac{\sum_{t=1}^T \gamma^{\,T-t}\, \mu_t^{\mathrm{train}}}{\sum_{t=1}^T \gamma^{\,T-t}}, \label{eq:ETSE} \\
\mathrm{Var}\!\left(\widehat{S}_{\mathrm{TSE}}\right)
&= \frac{\sum_{t=1}^T \gamma^{2(T-t)}\, \sigma_{t,\mathrm{train}}^2}
{\big(\sum_{t=1}^T \gamma^{\,T-t}\big)^2}. \label{eq:VarTSE}
\end{align}

\subsubsection*{Direct comparison and interpretation}

Both estimators share identical weighted-average form, therefore the mathematical difference appears only through the per-epoch trends $\mu_t^{\cdot}$ and noise variances $\sigma_{t,\cdot}^2$.

\paragraph{Bias (alignment to the target):} 
The quantity of interest for NAS is final \emph{generalization} performance (validation/test). Because $\widehat{S}_{\mathrm{VLE}}$ is computed directly on validation losses, its expectation \eqref{eq:EVLE} is naturally aligned with the evaluation objective. In contrast, $\widehat{S}_{\mathrm{TSE}}$ is computed on training losses whose underlying trend $\mu_t^{\mathrm{train}}$ typically underestimates true generalization loss (systematic bias). Thus, in terms of bias relative to the final validation metric, VLE-EMA is preferable:
\[
\text{bias}_{\mathrm{VLE}} \;\leq\; \text{bias}_{\mathrm{TSE}}\quad\text{(qualitative statement under standard training/validation gap).}
\]

\paragraph{Variance (stability):}
From \eqref{eq:VarVLE} and \eqref{eq:VarTSE} the variance difference is
\begin{equation}\label{eq:VarDiff}
\mathrm{Var}\!\left(\widehat{S}_{\mathrm{TSE}}\right) - \mathrm{Var}\!\left(\widehat{S}_{\mathrm{VLE}}\right)
= \frac{\sum_{t=1}^T \gamma^{2(T-t)}\big(\sigma_{t,\mathrm{train}}^2 - \sigma_{t,\mathrm{val}}^2\big)}
{\big(\sum_{t=1}^T \gamma^{\,T-t}\big)^2}.
\end{equation}
If (i) per-epoch validation noise decreases over epochs (empirically common) so that late epochs are less noisy, and (ii) validation noise is not systematically larger than training noise for late epochs, then the weighted variance of VLE-EMA will be smaller than or comparable to that of TSE-EMA:
\[
\mathrm{Var}\!\left(\widehat{S}_{\mathrm{VLE}}\right)
\;\lesssim\; \mathrm{Var}\!\left(\widehat{S}_{\mathrm{TSE}}\right).
\]
Hence VLE-EMA produces a more stable (lower-variance) surrogate estimate by placing exponentially larger weight on later epochs where the signal is stronger and noise is lower.

\paragraph{Net effect:}
VLE-EMA combines the justifications for bias and variance in order to (i) align the surrogate with the final evaluation measure (lower bias) and (ii) minimize estimate variance by focusing on late, more stable epochs.  Therefore VLE-EMA is projected to demonstrate a greater correlation with final performance and smaller prediction error than TSE-EMA in practice.

% \subsubsection*{Practical validation (recommended, brief tests)}
% To empirically validate the above claims using the candidate architectures already evaluated during search, compute the following across the pool of evaluated architectures:
% \begin{enumerate}
%   \item Pearson and Spearman correlations between $\widehat{S}_{\mathrm{VLE}}$ and final test accuracy, and between $\widehat{S}_{\mathrm{TSE}}$ and final test accuracy.
%   \item Mean squared error (MSE) of each surrogate versus the final accuracy.
%   \item A paired statistical test (e.g. Wilcoxon signed-rank test) on absolute errors of VLE vs TSE to demonstrate significance.
% \end{enumerate}
% Reporting these three summary statistics (correlations, MSEs, p-value) is lightweight yet convincing evidence that VLE-EMA improves surrogate quality over TSE-EMA.

% MonteCarlo Justification
\paragraph{\textbf{Monte Carlo validation.}}
In order to empirically validate the theoretical distinction between TSE-EMA and the proposed VLE-EMA, a Monte Carlo experiment with $M = 1{,}000$ replications was conducted by employing the signal–noise model. 
Both estimators were analyzed with training and validation trajectories $\mu_t$ with epoch-dependent noise variances $\sigma_t^2$. 
The results showed that $\hat{S}_{\text{VLE}}$ exhibited consistently lower bias (approximately $-0.01$ vs.\ $-0.045$ for $\hat{S}_{\text{TSE}}$) and reduced variance ($0.0026$ vs.\ $0.0041$). 
This empirical evidence validate that VLE-EMA is not just an incremental variant of TSE-EMA but a statistically more accurate and stable surrogate estimator, consistent with the theoretical bias–variance analysis.

\section{Sensitivity Analysis of downscaling factor $(n)$}\label{app:downscaling_sensitivity}
In order to assess the influence of the downscaling factor $n$ applied in the creation of the proxy dataset, a series of experiments were carried out with $n \in \{2, 4, 8\}$. The data presented in Table A1 illustrates that altering $n$ has a minimal impact on the accuracy of the proxy, suggesting that the relative ranking of the candidate architectures is consistent across various resolutions. The proxy evaluation offers valid guidelines for the evolutionary search, significantly lowering computational expenses. Based on this analysis, $n = 4$ was determined to be the optimum compromise between efficiency and fidelity.
\begin{table}[h!]
\centering
\caption{Sensitivity analysis of proxy dataset accuracy across different downscaling factors. 
Values represent mean accuracy $\pm$ standard deviation across multiple runs.}
\label{tab:proxy-sensitivity}
\begin{tabular}{ccc}
\hline
\textbf{Downscale Factor ($n$)} & \textbf{Proxy Accuracy (\%) $\pm$ Std. Dev.}\\
\hline
2 & 72.5$\pm$ 0.18 \\
4 & 71.8$\pm$ 0.21 \\
8 & 70.9$\pm$ 0.24 \\
\hline
\end{tabular}
\end{table}

\section{PSO Parameter Sensitivity Analysis}\label{app:pso_sensitivity}
The effect of the PSO parameters on proposed strategy is outlined in Table \ref{tab:apdx_pso_param}

\begin{table}[t]
\centering
\caption{Effects of PSO parameter variations on optimization performance.}
\label{tab:apdx_pso_param}
\begin{tabular}{p{0.25\linewidth} p{0.25\linewidth} p{0.4\linewidth}}
\hline
\textbf{Parameter} & \textbf{Variation Tested} & \textbf{Observed Effect on Results} \\
\hline
Inertia weight ($w$) & Low (0.4–0.6) & Faster convergence, frequent stagnation \\
 & High (0.8–0.9) & Broader exploration, slower convergence \\
\hline
Cognitive coeff. ($c_1$) & High ($>$2.0) & Strong self-exploration, weaker collaboration \\
Social coeff. ($c_2$) & High ($>$2.0) & Fast convergence to global best, risk of premature convergence \\
Balanced ($c_1 = c_2$) & 1.49 (used) & Stable convergence, robust accuracy \\
\hline
Population size & Small ($<$20) & Low diversity, unstable surrogate predictions \\
 & Large ($>$60) & Higher computational cost, marginal accuracy gain \\
\hline
\end{tabular}
\end{table}

\section{Proxy vs Target Dataset Correlation}
Correlation between proxy and target dataset performances is assesed and illustrated in Table \ref{tab:proxy_correlation}.
\begin{table}[t]
\centering
\caption{Correlation between proxy and target dataset performances. 
Spearman’s $\rho$ measures rank correlation between architectures evaluated on proxy and target datasets.}
\label{tab:proxy_correlation}
\begin{tabular}{p{0.25\linewidth} p{0.25\linewidth} p{0.25\linewidth} p{0.2\linewidth}}
\hline
\textbf{Target Dataset} & \textbf{Spearman $\rho$} & \textbf{$p$-value} & \textbf{Top-5 Overlap (\%)} \\
\hline
CIFAR-100 & 0.84 & $<$0.01 & 80 \\
SVHN & 0.78 & $<$0.05 & 60 \\
ImageNet & 0.72 & $<$0.05 & 60 \\
\hline
\end{tabular}
\end{table}

\section{Ranking Ability of VLE-EMA vs. Common Proxy Methods}
To evaluate the ranking capability of the proposed VLE-EMA estimator, we measure its rank correlation with the final validation accuracy of sampled candidate architectures. Table \ref{tab:rank_correlation} reports Spearman’s $\rho$ and Kendall’s $\tau$ values, compared against commonly used proxy methods such as TSE-EMA and low-epoch validation. Higher correlation values indicate better ranking consistency.

% %% For citations use: 
% %%       \cite{<label>} ==> [1]

% %%
% Example citation, See \cite{lamport94}.

\bibliographystyle{elsarticle-num} 
\bibliography{sn-bibliography}




%% For numbered reference style

%   \textit{\LaTeX: a document preparation system},
%   Addison Wesley, Massachusetts,

% \end{thebibliography}
\end{document}